\documentclass{article}

\PassOptionsToPackage{numbers, compress}{natbib}

\usepackage[preprint]{neurips_2026}

\usepackage[utf8]{inputenc} 
\usepackage[T1]{fontenc}    
\usepackage{hyperref}       
\usepackage{url}            
\usepackage{booktabs}       
\usepackage{amsfonts}       
\usepackage{nicefrac}       
\usepackage{microtype}      
\usepackage{xcolor}         
\usepackage{pifont}
\usepackage[table]{xcolor}
\usepackage{booktabs}
\usepackage{multirow}
\usepackage{caption}
\usepackage{wrapfig}
\usepackage{graphicx}
\usepackage{soul}
\usepackage{ulem}
\usepackage{enumitem}
\usepackage[T1]{fontenc}
\usepackage{comment}
\usepackage{subcaption}
\usepackage{amsmath}
\usepackage{xspace}
\usepackage{fontawesome5}

\definecolor{rowgray}{RGB}{245,245,250}
\definecolor{rowgray}{RGB}{245,245,250}
\definecolor{metricblue}{RGB}{220,230,242}
\definecolor{metricgreen}{RGB}{226,239,218}
\definecolor{rowblue}{RGB}{214, 227, 242} 
\definecolor{gtgreen}{HTML}{3F7F2F}
\definecolor{darkred}{HTML}{8B1E1E}

\title{EpiGraph: Building Generalists for Evidence-Intensive Epilepsy Reasoning in the Wild}

\author{%
  Yuyang Dai$^{1}$ \quad
  Zheng Chen$^{2}$\textsuperscript{\dag} \quad
  Jathurshan Pradeepkumar$^{3}$ \quad
  Yasuko Matsubara$^{2}$ \\
  \textbf{Jimeng Sun$^{3}$} \quad
  \textbf{Yasushi Sakurai$^{2}$} \quad
  \textbf{Yushun Dong$^{1}$}\textsuperscript{\dag} \\[4pt]
  $^{1}$Florida State University \quad
  $^{2}$The University of Osaka \quad
  $^{3}$University of Illinois Urbana-Champaign
}

\begin{document}
\maketitle
\footnotetext{\textsuperscript{\dag}Corresponding authors:
\texttt{chenz@sanken.osaka-u.ac.jp},
\texttt{yd24f@fsu.edu}}

\begin{abstract}

Epilepsy diagnosis and treatment require evidence-intensive reasoning across heterogeneous clinical knowledge, including biosignal patterns, genetic mechanisms, pharmacogenomics, treatment strategies, and patient outcomes. 
In this work, we present \textsc{EpiGraph}, a large-scale epilepsy knowledge graph and benchmark for evaluating knowledge-augmented clinical reasoning. \textsc{EpiGraph} integrates 48,166 peer-reviewed papers and seven clinical resources into a heterogeneous graph containing 24,324 entities and 32,009 evidence-grounded triplets across five clinical layers. Built upon this graph, \textsc{EpiBench} defines five clinically motivated tasks spanning clinical decision-making, EEG report generation, pharmacogenomic precision medicine, treatment recommendation, and deep research planning.
We evaluate six LLMs under both standard and Graph-RAG settings. Results show that integrating \textsc{EpiGraph} consistently improves performance across all tasks, with the largest gains observed in pharmacogenomic reasoning (+30--41\%). Our findings demonstrate that structured epilepsy knowledge substantially enhances evidence-grounded clinical reasoning and provides a practical benchmark framework for evaluating knowledge-augmented LLMs in real-world neurological settings.

\begin{center}
\small
{\large\faGithub}~\href{https://github.com/LabRAI/EpiGraph}{\texttt{https://github.com/LabRAI/EpiGraph}}
\qquad
{\large\faGlobe}~\href{https://labrai.github.io/EpiGraph/}{\texttt{https://labrai.github.io/EpiGraph/}}
\qquad
{\large\faDatabase}~\href{https://huggingface.co/RAI-Lab/EpiGraph}{\texttt{https://huggingface.co/RAI-Lab/EpiGraph}}
\end{center}

\end{abstract}

\section{Introduction}

Epilepsy is a long-standing challenge in neuroscience and clinical medicine, affecting over 50 million individuals \cite{kotoge2025evobrain}.
Although many disease mechanisms can lead to epilepsy, accurate diagnosis remains challenging \cite{kwan2011drug}.
To answer diagnostic questions, clinicians must integrate multiple sources of evidence, including seizure semiology, electrophysiological signals, and patient history \cite{SODorAAAI2025}. 
This process requires extensive training to construct a comprehensive knowledge framework.
Moreover, the etiology of epilepsy remains unknown in approximately 50\% of cases, and understanding disease progression requires not only identifying subtle symptoms, but also determining which causal genes are associated with the disease or characterizing how patients respond to treatments \cite{kuo2024pharmacogenetics}.
This requires integrating heterogeneous knowledge across biological, clinical, and pharmacological perspectives, as well as reasoning evidence from a large body of literature. 
\textit{Precise curation and reasoning of such diverse evidence} is therefore crucial for both clinical practice and scientific discovery.

To support such evidence-intensive investigation, heterogeneous biomedical knowledge must be represented in a structured and integrative form. 
Knowledge graphs (KGs) provide a natural framework for this purpose, organizing entities and their relationships across multiple domains and enabling multi-hop reasoning~\cite{cui2025review}.
Several biomedical KGs have been proposed across diverse domains such as precision medicine~\cite{chandak2023building}, drug repurposing~\cite{lu2025biomedical}, oncology~\cite{autoRD}, and disease–gene–drug modeling~\cite{bonner2022review}.
However, prior works either remain disease-agnostic and lack epilepsy-specific knowledge bases~\cite{bodenreider2004umls}, or are designed for semantic annotation, which focuses on identifying medical terms in text (e.g., symptoms, seizure types, and electrophysiological patterns) and converting them into a consistent vocabulary~\cite{sahoo2014epilepsy, sargsyan2023epilepsy}. 
Therefore, there remains a gap in developing a KG that comprehensively captures epilepsy-specific knowledge and enables interpretation of complex underlying disease mechanisms.

\begin{figure*}[t]
    \centering
    \includegraphics[width=0.93\linewidth]{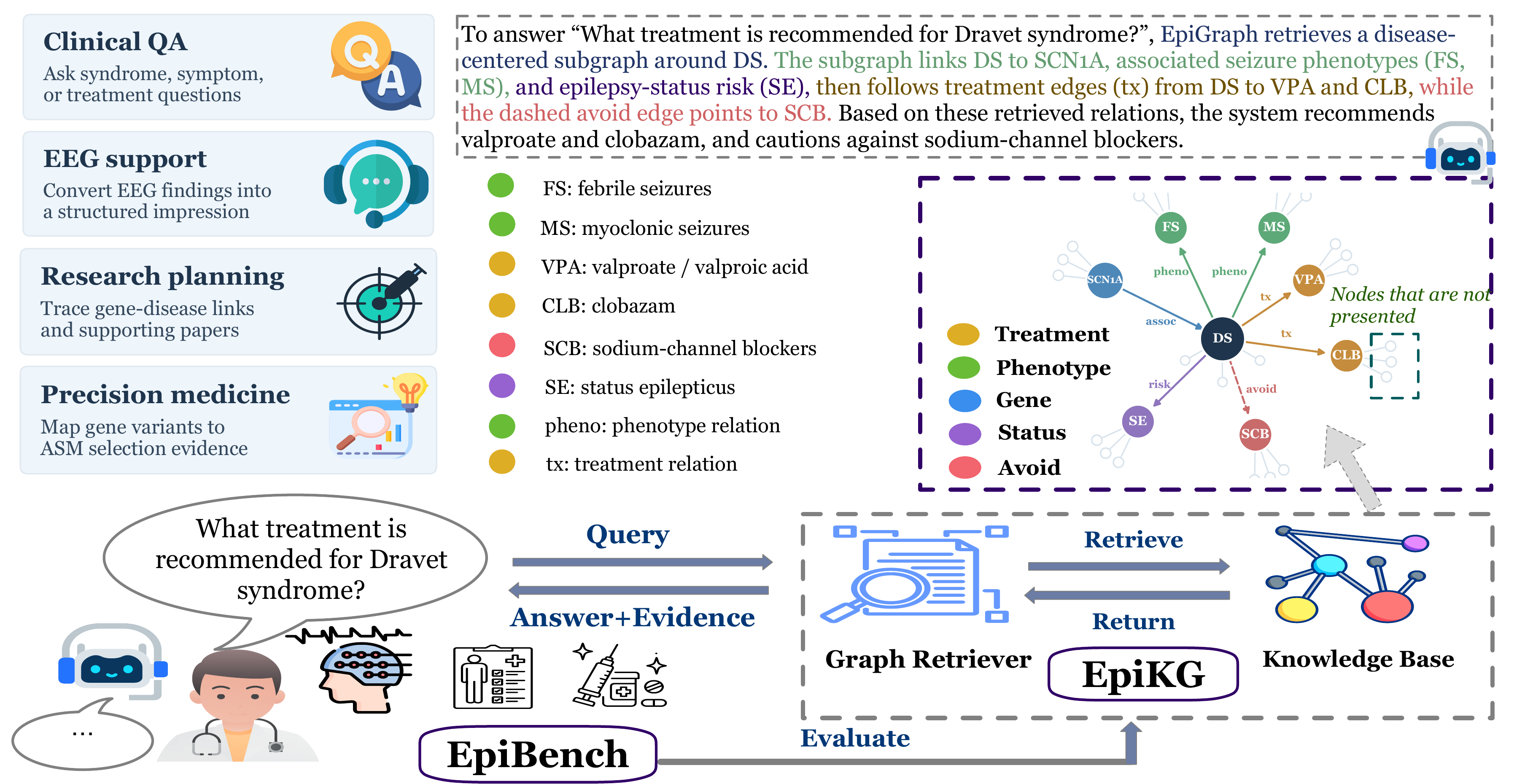}
    \caption{\textbf{Overview of the \textsc{\textbf{EpiGraph}} framework}, comprising two core components: \textsc{EpiKG}, an epilepsy knowledge graph, and \textsc{EpiBench}, a multi-task evaluation benchmark. Given a clinical query (e.g., ``What treatment is recommended for Dravet syndrome?''), the Graph Retriever queries \textsc{EpiKG} to extract a subgraph linking the syndrome to associated phenotypes, genes, treatments, and contraindications. The retrieved
reasoning paths are returned to the LLM to generate a grounded
answer with supporting evidence, which is then evaluated by
\textsc{EpiBench} across five clinical tasks.}
        \vspace{-0.6cm}
    \label{fig:EpiGraph_overview}

\end{figure*}

Recently, large language models (LLMs) have demonstrated strong capabilities for advancing evidence curation in scientific research, leveraging both internal knowledge corpora and the ability to interpret external scientific publications \cite{singhal2023large, kung2023performance, nori2023capabilities, shao2025llm}. 
Recent work further combines KGs with LLMs for clinical reasoning~\cite{gao2025drknows,rezaei2025amgrag,chen2025medreason, li2026llm}. 
A growing body of benchmarks has been proposed to evaluate LLMs on complex biomedical question answering and multi-step reasoning~\cite{al2025artificial, lucas2024artificial,ma2024clibench,arora2025healthbench,fan2025diagnosisarena,xiong2024benchmarking}.
However, existing benchmarks often focus on contrived tasks within narrow domains that are not reflective of epilepsy.
More importantly, epilepsy research extends beyond isolated clinical question answering, requiring integration across biomarker discovery, disease mechanisms, treatment strategies, and broader clinical practice \cite{lincanICHI}. 
Therefore, a critical gap exists in datasets to evaluate whether LLMs can perform expert-level evidence curation and reasoning in realistic epilepsy settings.

\textbf{Present work.} 
In this work, we identify two key limitations in prior KGs and evaluation for epilepsy, and introduce \textsc{EpiGraph}, a unified framework for evidence-intensive reasoning in epilepsy, as shown in Figure \ref{fig:EpiGraph_overview}, which consists of two components. 
\ding{182} \textsc{EpikG}, a large-scale epilepsy KG constructed via a structured evidence-to-graph pipeline with three stages. 
First, we define the graph schema based on seven authoritative resources (e.g., NCBI-MeSH \cite{nlm_mesh}), organizing entities into five layers (genes, phenotypes, syndromes, treatments, and outcomes) and defining 1,370 cross-layer relation types (e.g., \textit{caused by gene}, \textit{treated with}).
This ensures that the graph is grounded in clinically validated resources. 
We then collect large-scale evidence over 120,000 PubMed papers and applying a two-stage screening process, where an LLM-assisted classifier filters candidates and domain experts adjudicate borderline cases, yielding 48,166 papers for evidence extraction. 
We finally map the extracted evidence into the predefined graph schema, aligning relations to predefined types, and validating triplets. 
The resulting graph contains 24,324 entities and 32,009 triplets, including 14,576 cross-layer connections, enabling evidence-grounded, multi-hop reasoning in epilepsy. 
\ding{183} \textsc{EpiBench}, a multi-task benchmark for evaluating LLM reasoning and the effectiveness of \textsc{EpikG} in epileptology.
Our evaluation covers five tasks, including biomedical question answering, as well as a practical clinical task that supports clinicians in EEG interpretation and report generation.
To construct the benchmark, we curate evidence from 1,700 research papers involving over 7,000 clinical cases, and transform them into structured evaluation formats. 
This results in 6,199 QA pairs, 151 pharmacogenomic multiple-choice questions, 472 treatment recommendation tasks, and 163 research planning cases. 
For the practical EEG-to-report generation task, LLMs take EEG inputs and incorporate external knowledge in \textsc{EpikG} to interpret the data and generate an evidence-grounded report.

\textbf{Our contributions} are as follows: \\
\ding{113} \textit{A New Epilepsy Knowledge Graph.} We propose the first large-scale, multi-hop relational knowledge graph dedicated to epileptology, spanning phenotypes, genetic mechanisms, and treatments.\\
\ding{113} \textit{A Comprehensive Benchmark Dataset.} We propose a publicly released, modular benchmark spanning five clinical reasoning tasks to support plug-and-play evaluation of any LLM or retriever.\\
\ding{113} \textit{A Clinically Grounded Evaluation with Practical Impact.} 
We validate on decades of real-world Harvard clinical EEG notes, grounding evaluation in authentic clinical practice and demonstrating that graph-augmented LLMs can meaningfully support AI-assisted epilepsy care.

\section{\textbf{\textsc{EpiKG}}: Epilepsy Knowledge Graph}
\label{sec:EpiKG}

\subsection{Problem Formulation}

We define \textsc{EpiKG} as a multi-relational heterogeneous knowledge graph $\mathcal{G} = (\mathcal{V}, \mathcal{E})$, where $\mathcal{V}$ denotes the set of epilepsy-related entities and $\mathcal{E}$ denotes the set of typed relations between entities. 
Each entity belongs to one of five layers, including \textit{genes, phenotypes, syndromes, treatments, and outcomes}. 
We collect a set of associated scientific documents $\mathcal{D} = \bigcup_{v \in \mathcal{V}} \mathcal{D}_v$, to extract various scientific findings or clinical observations for representing $v$, via a mapping function $\mathcal{P}$.
Each relation is supported by evidence collected from clinical guidelines,
where $\mathcal{D}_v$ denotes the collection of scientific documents associated with entity $v$.
Therefore, each fact in $\mathcal{G}$ is represented as a triplet $\langle h, r, t \rangle$, where $h$ and $t$ are entities and $r$ denotes the relation type between them. 
Cross-layer triplets connect entities across distinct clinical layers, enabling multi-hop reasoning for epilepsy diagnosis, treatment, and discovery.


\subsection{Data Collection and Processing}

\textbf{Data Source.}
To construct the comprehensive entities $\mathcal{V}$ in \textsc{EpiKG}, we collect large-scale epilepsy-related scientific literature $\mathcal{D}$ from PubMed/PMC~\cite{nlm_mesh}. Using a comprehensive MeSH-based retrieval strategy covering epilepsy, seizure, epileptic encephalopathy, antiseizure medication, EEG, and pharmacogenomics, we retrieve over 120,000 candidate publications indexed between 1990 and 2024. These publications serve as the primary evidence source for extracting cross-layer clinical relations across genes, phenotypes, syndromes, treatments, and outcomes.\\
\textbf{Ontology and Guideline Resources.}
To ensure clinically grounded and comprehensive relational edges $\mathcal{E}$, the schema of \textsc{EpiKG} is constructed from seven authoritative clinical resources, including ILAE 2022~\cite{fisher2014ilae}, MeSH~\cite{nlm_mesh}, OMIM~\cite{hamosh2005omim}, ChEBI~\cite{hastings2016chebi}, HPO~\cite{kohler2021human}, AES 2024~\cite{aes2024}, and UMLS~\cite{bodenreider2004umls}. 
These resources provides curated term lists, concept hierarchies, and cross-resource identifiers. 
ILAE 2022 provides the authoritative classification of epilepsy syndromes and seizure types. OMIM provides gene--disease associations as the primary source for Gene entities. 
ChEBI provides standardized chemical identifiers for antiseizure medications in Treatment. HPO provides standardized phenotype vocabulary for seizure types and developmental outcomes. 
AES 2024 provides evidence-based treatment recommendations grounding Treatment relations. UMLS serves as the cross-ontology linking hub, integrating 200+ controlled vocabularies to enable synonym resolution and entity alignment.\\
\noindent \textbf{Preprocessing.}
We preprocess both the literature corpus and ontology resources through a two-stage pipeline. 
In the \textit{automated filtering stage}, an LLM-based classifier is applied to titles and abstracts to exclude: (i)~papers not primarily focused on epilepsy or seizure disorders; (ii)~purely animal or in vitro studies without clinical relevance; (iii)~case reports with limited generalizability; (iv)~non-English publications; and (v)~duplicate records. In the \textit{domain expert review stage}, borderline cases are adjudicated using established epilepsy systematic-review criteria~\cite{beghi2019global, kwan2011drug}, retaining studies reporting original findings on syndrome classification, genetic etiology, ASM efficacy and safety, EEG biomarkers, pharmacogenomics, and treatment outcomes. 
Concurrently, ontology resources are normalized by resolving cross-resource identifier conflicts through UMLS CUI mapping and curating over 100 term lists covering entity aliases, abbreviations, and relation patterns across all layers. This process yields a corpus of 48,166 papers and a validated entity vocabulary for graph construction.


\begin{figure*}[t]
    \centering
    \includegraphics[width=\textwidth]{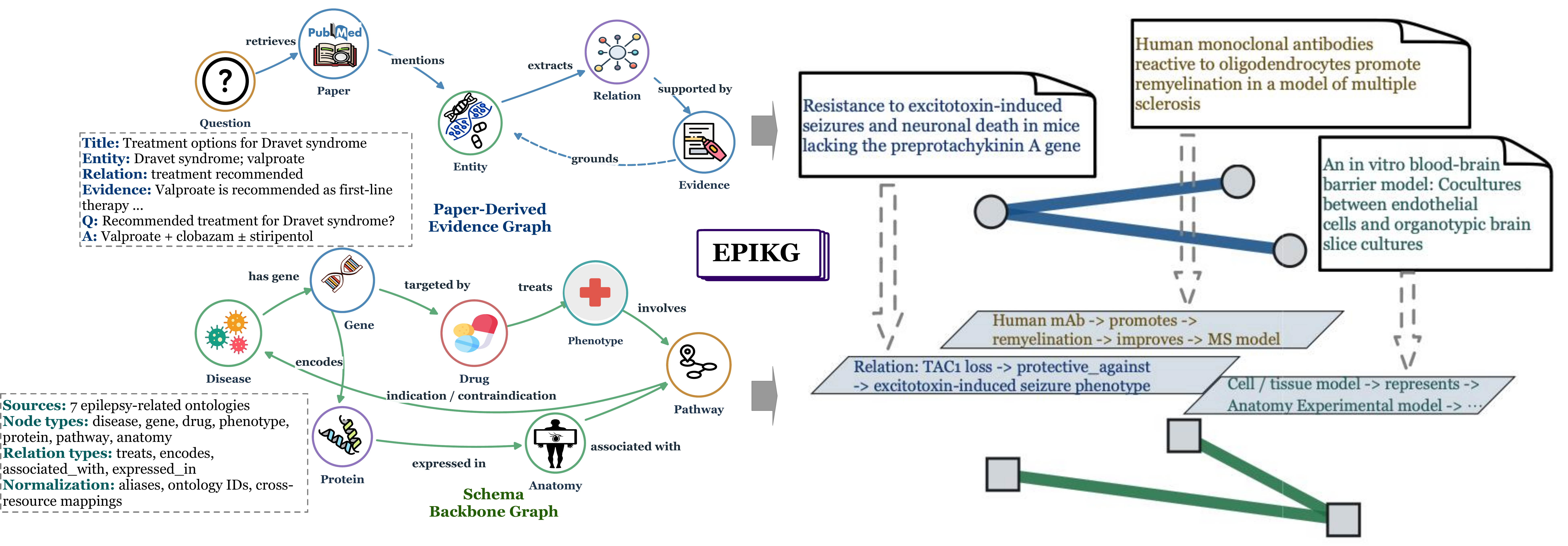}
    \caption{
Pipeline overview of \textsc{EpiKG}, comprising
two components.
\textit{Left:} \textcolor[HTML]{17375E}{paper-derived evidence graph} is processed through an extraction pipeline that identifies entities, relations, and supporting evidence, mapped into
\textcolor[HTML]{1F5C2E}{relation graph}.
\textit{Right:} An example of how \textsc{EpiKG} grounds the \textcolor[HTML]{17375E}{paper-derived evidence graph}: given the
query paper \textcolor[HTML]{17375E}{``Resistance
to excitotoxin-induced seizures...''},
\textsc{EpiKG} retrieves the supporting reasoning
path, linking the retrieved evidence back to the \textcolor[HTML]{1F5C2E}{relation graph}.}
        \vspace{-0.4cm}
    \label{fig:EpiKG_overview}
\end{figure*}


\subsection{Knowledge Graph Construction}

\paragraph{Entity Extraction.}
We extract five layers (\textbf{L}) of entities for
\textsc{EpiKG}, each corresponding to one clinical layer.
For example, given the sentence \textit{``Patients with Dravet Syndrome carrying SCN1A loss-of-function variants showed seizure freedom after treatment with Stiripentol combined with Valproate''}, the pipeline extracts:
\textit{Dravet Syndrome} (L1), \textit{SCN1A} (L3),
\textit{Stiripentol} and \textit{Valproate} (L4), and
\textit{Seizure Freedom} (L5), which are subsequently linked as typed relations in \textsc{EpiKG}
(Figure~\ref{fig:EpiKG_overview}).
Therefore, the five layers are as follows:
\\
\textbf{- L1 Syndrome.} This layer includes epilepsy syndromes and seizure types following the ILAE 2022 taxonomy~\cite{fisher2014ilae}, including Dravet Syndrome, Lennox-Gastaut Syndrome, and West Syndrome. Entities are extracted from ILAE 2022 and MeSH~\cite{nlm_mesh}, and normalized through UMLS CUI mapping.
\\
\textbf{- L2 Diagnostic.} Diagnostic entities capture EEG patterns, neuroimaging findings, and biomarkers, such as Spike-Wave Discharge, Hypsarrhythmia, and MRI T2 Signal Abnormality. Entities are extracted from MeSH and HPO~\cite{kohler2021human}. 
\\
\textbf{- L3 Gene.} 
Gene entities encode causal genetic factors associated with epilepsy, including genes and pathogenic variants. Examples include SCN1A, KCNQ2, TSC1, and CDKL5. 
Entities are extracted from OMIM gene--disease association~\cite{hamosh2005omim} and normalized using HGNC identifiers. 
\\
\textbf{- L4 Treatment.} Treatment entities cover antiseizure medications, surgical procedures, and neuromodulation therapies, including Valproate, Stiripentol, Vagus Nerve Stimulation, and Ketogenic Diet. Entities are extracted from ChEBI~\cite{hastings2016chebi} chemical identifiers and AES 2024~\cite{aes2024} guideline-listed interventions, with abbreviation normalization handled through curated aliases. 
\\
\textbf{- L5 Outcome.} Outcome entities represent clinical endpoints such as seizure control, adverse effects, and developmental outcomes, including Seizure Freedom, Drug Resistance, Cognitive Impairment, and SUDEP. Entities are extracted from HPO~\cite{kohler2021human} and MeSH.

\vspace{-0.3cm}
\paragraph{Relation Construction.}
Relations are extracted through two complementary pipelines.
(i) Rule-based extraction applies ontology-derived pattern
matching to sentences containing co-occurring entity pairs.
For example, given \textit{``Valproate is
recommended as first-line treatment for Dravet Syndrome but
should be avoided in patients with SCN1A gain-of-function
variants''}, the pipeline matches the pattern
\texttt{[Treatment] + \{recommended for / first-line\} +
[Syndrome]} to extract the triplet
(Valproate, \texttt{treats}, Dravet Syndrome),
and \texttt{[Treatment] + \{avoid / contraindicated\} +
[Gene]} to extract
(Valproate, \texttt{contraindicated\_with},
SCN1A). 
(ii) LLM-based extraction employs MiniMax-Text-01~\cite{minimax2025minimax01} on full-text articles with structured prompts specifying entity layers and relation types, enabling extraction of novel associations beyond rule-based coverage. 
\\
\textbf{- Etiology Relations.} 
These relations connect genetic factors with epilepsy syndromes and encode causal disease mechanisms, e.g., $\langle \textit{SCN1A}, \textsc{caused\_by\_gene}, \textit{Dravet Syndrome} \rangle$.
\\
\textbf{- Diagnostic Relations.} 
These relations connect syndromes with clinical manifestations and diagnostic findings, e.g., $\langle \textit{West Syndrome}, \textsc{characterized\_by}, \textit{Hypsarrhythmia} \rangle$.
\\
\textbf{- Treatment Relations.} 
These relations encode therapeutic associations between syndromes, genes, and interventions, e.g., $\langle \textit{Dravet Syndrome}, \textsc{treated\_with}, \textit{Stiripentol} \rangle$.
\\
\textbf{- Outcome Relations.} 
These relations capture prognostic outcomes and treatment effects, e.g., $\langle \textit{Valproate}, \textsc{causes\_teratogenicity}, \textit{Neural Tube Defects} \rangle$.

\subsection{Graph Mapping and EpiKG Statistics}

After entity extraction and relation construction, raw literature evidence must be mapped to the curated entity vocabulary and validated before integration into \textsc{EpiKG}. This process consists of three stages.
\textit{First, Entity Normalization.}
Literature mentions are normalized to canonical entity identifiers through a four-stage pipeline: (i) a exact match maps mentions to curated term lists; (ii) an alias match resolves abbreviations and synonyms (\textit{e.g.}, VPA $\rightarrow$ Valproate, Nav1.1 $\rightarrow$ SCN1A); (iii) a semantic match uses sentence-transformer embeddings~\cite{SBERT} and cosine similarity to retrieve the closest entity for unresolved mentions; and (iv) a quality control verifies consistency through UMLS CUI alignment across ontologies.
\textit{Second, Relation Type Matching.}
Extracted relation candidates are aligned to the 1,370 predefined cross-layer relation types. Rule-based candidates are matched through ontology-derived templates, while LLM-extracted relations are normalized using ontology-grounded semantic matching. Low-confidence candidates are flagged for expert review.
\textit{Finally, Triplet Validation.}
Validated triplets are deduplicated, assigned to entity layers, and annotated with paper count as a proxy for evidential strength. 
Low-evidence triplets supported by fewer than two independent sources are retained but flagged accordingly. 

The resulting \textsc{EpiKG} contains 32,009 triplets across 24,324 unique entities and 1,370 cross-layer relation types. 
Rule-based extraction contributed 9,670 triplets (30.2\%) and LLM-based extraction 22,339 triplets (69.8\%). Cross-layer triplets number 14,576 (45.5\%), covering all pairwise layer combinations. The densest cross-layer connections are between L1~Syndrome and L4~Treatment (3,217 triplets) and between L3~Gene and L1~Syndrome (2,845 triplets), capturing the gene--syndrome--treatment reasoning chains central to clinical decision-making. The median paper count per triplet is 3 (IQR: 1--8), and 4,612 triplets are supported by at least 10 independent publications. 
\section{EpiBench}
\label{sec:epibench}

\subsection{Problem Formulation}
\label{sec:problem_formulation}

We formulate \textsc{EpiBench} as a unified benchmark for evaluating evidence-intensive reasoning in epilepsy across both clinical and scientific settings. 
Each task is defined as:
\begin{equation}
    \hat{y}_i = f_{\mathrm{LM}}(x_i, \mathcal{E}_i, \mathcal{C}_i),
    \label{eq:general}
\end{equation}
where $x_i$ denotes the primary task input, including clinical cases, EEG recordings, or scientific documents; $\mathcal{E}_i$ denotes the evidence context retrieved from \textsc{EpiKG} through Graph-RAG; $\mathcal{C}_i$ denotes optional task-specific context such as candidate options, supporting entities, or document metadata; and $\hat{y}_i$ denotes the model prediction evaluated against the gold-standard output $y_i^*$. 

The benchmark covers diverse reasoning scenarios, including diagnosis-oriented question answering, pharmacogenomic treatment, biomarker discovery, research planning, and EEG-to-report generation.

\subsection{Task Design}
\label{sec:task_design}

Epilepsy management requires jointly reasoning over syndrome classification, seizure phenotype, genetic factors, treatment contraindications, and patient-specific context. 
Accordingly, \textsc{EpiBench} aims to evaluate reasoning, including clinical decision-making from phenotypic evidence, biomarker-driven precision medicine, treatment recommendations, and deep research planning from scientific literature. 
Beyond conventional QA settings, we further introduce a practical EEG-to-report generation task that reflects real-world clinical workflow \cite{tatum2016american,pradeepkumar2026neural}. 
We investigate whether LLMs, through interaction with \textsc{EpiGraph}, can leverage external epilepsy knowledge to support deeper EEG interpretation, clinical reasoning, and automatic neurologist-style report generation. The tasks are as follows.


\textbf{Task 1: Clinical Decision Accuracy (CDA).}
This task evaluates epilepsy-specific clinical reasoning using recently published papers excluded from \textsc{EpiKG} to prevent knowledge leakage~\cite{singhal2023large,singhal2025toward}. The benchmark includes 1,000 multiple-choice questions and 5,199 open-ended questions, covering syndrome diagnosis, EEG interpretation, treatment selection, and phenotype reasoning.
For example, given the question ``A child with febrile seizures and \textit{SCN1A} mutation is diagnosed with Dravet syndrome. What is the first-line treatment?'', the model must select
corresponding treatment from different options.
Ground-truth answers are derived from expert-curated source literature.
\\
\textbf{Task 2: Clinical Report Generation (CRG).}
This task focuses on clinical impression generation from patient information and EEG descriptions (i.e., a summary of EEG signals). 
Given these inputs, LLMs must identify important findings, including characteristic EEG waveforms, dominant frequency patterns, and their associations with disease states, and then generate an impression for real-time patient documentation. 
Different from conventional QA benchmarks, CRG evaluates whether LLMs can perform evidence-grounded clinical reasoning or through interaction with \textsc{EpiKG}, integrating EEG findings with syndrome knowledge and evidence. 
\\
\textbf{Task 3: Biomarker-Driven Precision Medicine (BPM).}
This task evaluates whether LLMs can select appropriate antiseizure medications (ASMs) from genetic variants and patient phenotypes. It requires multi-step reasoning over gene function, disease mechanisms, drug targets, and contraindication evidence~\cite{kuo2024pharmacogenetics}. The benchmark contains 151 multiple-choice questions constructed from CPIC and ILAE 2022 guidelines.
For example, given a patient with a \textit{TSC2} variant and refractory focal seizures, the model should select \textit{Everolimus} over \textit{Carbamazepine} by reasoning through the path:
$
\textit{TSC2}
\rightarrow
\textit{mTOR Pathway}
\rightarrow
\textit{Everolimus},
$
while recognizing \textit{Carbamazepine} as the contraindicated.
\\
\textbf{Task 4: Treatment Recommendation (TR).}
This task evaluates whether LLMs can recommend guideline-consistent therapies under patient-specific constraints~\cite{glauser2006ilae}. Questions are constructed from neurology subsets of \textit{MedQA-USMLE}~\cite{jin2021disease} and \textit{MMLU Professional Medicine}~\cite{hendrycks2020measuring}, covering 472 treatment recommendation cases in total.
For example, given ``a woman of childbearing age with juvenile myoclonic epilepsy'', the model should recommend \textit{Levetiracetam} instead of \textit{Valproate}, while recognizing the risk of \textit{Valproate}.
\\
\textbf{Task 5: Deep Research Planning (DRP).}
This task evaluates whether LLMs can perform scientific reasoning from epilepsy literature by proposing feasible research directions~\cite{wang2025survey}. 
The benchmark is constructed from 163 PMC epilepsy papers, with expert annotations available for 30 papers.
For example, given a paper on \textit{KCNQ2}-related neonatal epilepsy, the model should identify unresolved questions regarding long-term neurodevelopmental outcomes and propose a longitudinal cohort study grounded in prior evidence from \textsc{EpiKG}. 
The generated plan should specify meaningful endpoints, such as neurodevelopmental outcomes.

\textbf{Evaluation and Metrics. }
We evaluate \textsc{EpiBench} using both task accuracy and reasoning-oriented metrics. 
For \textbf{CDA}, we report Top-1 Accuracy for MCQs and ROUGE-L, BERTScore F1, and LLM-as-Judge for open-ended responses, evaluating both answer correctness and reasoning quality. 
For \textbf{CRG}, we use ROUGE-L and provide text alignment human evaluation.
For \textbf{BPM} and \textbf{TR}, we report Top-1 Accuracy, Guideline Concordance, and Drug Safety Score to evaluate treatment selection, safety, and alignment with clinical recommendations. TR additionally includes KG Evidence Coverage, measuring whether predictions are supported by retrieved evidence paths from \textsc{EpiKG}.
For \textbf{DRP}, we use ROUGE-L, BERTScore F1, Alignment Score, and LLM-as-Judge to evaluate scientific validity, coherence, and feasibility of generated research plans.

\section{Experiments}
\label{sec:results}

\noindent\textbf{Models.}
We evaluate six LLMs spanning closed- and
open-source families. \textit{Closed-source: }
GPT-4o~\cite{openai2024gpt4o},
Claude Sonnet 4~\cite{anthropic2024claude},
and Gemini 2.0 Flash~\cite{google2024gemini}.
\textit{Open-source:}
Llama-3.3-70B~\cite{grattafiori2024llama3},
Qwen2.5-72B~\cite{yang2024qwen25}, and
Mistral Small 3.1~\cite{mistral2025small}.
All six models are evaluated on Task 1,3,4,5.
For Task2 (\textbf{CRG}), due to dataset
usage restrictions, evaluation is on \textit{four
locally deployed smaller models:}
Gemma-3-4B~\cite{google2024gemma},
Llama-3.2-3B~\cite{grattafiori2024llama3},
MedGemma-4B~\cite{google2024medgemma}, and
Qwen3-4B~\cite{yang2024qwen25}.

\noindent\textbf{Prompting.}
We construct prompts for all tasks following three principles:
(1) chain-of-thought prompting~\cite{wei2022chain} to elicit
step-by-step clinical reasoning; (2) role-playing prompts
informing the model it is a skilled epileptologist; and
(3) full task context as defined in \S\ref{sec:task_design},
including the retrieved \textsc{EpiKG} subgraph serialised as
structured reasoning paths.
For MCQ tasks, temperature is set to 0.0; for generation
tasks, to 0.3. All closed-source models are accessed via the OpenRouter API; all open-source models are deployed locally under standardised inference settings to ensure reproducibility. Full details see Appendix~\ref{sec:app_prompts}.

\noindent\textbf{Evaluation.}
Standard metrics include Top-1 Accuracy (MCQ), ROUGE-L,
BERTScore F1, and LLM-as-Judge (GPT-4.1-mini, 1--5 scale).
Domain-specific metrics include Clinical NER F1, Hallucination
Rate (NLI-based), Guideline Concordance (ILAE 2022 / CPIC),
Drug Safety Score (contraindication avoidance), KG Evidence
Coverage, and Alignment Score. Full metric definitions see Appendix~\ref{sec:app_metrics}.

\begin{figure*}[t]
    \centering
    \includegraphics[width=\textwidth]{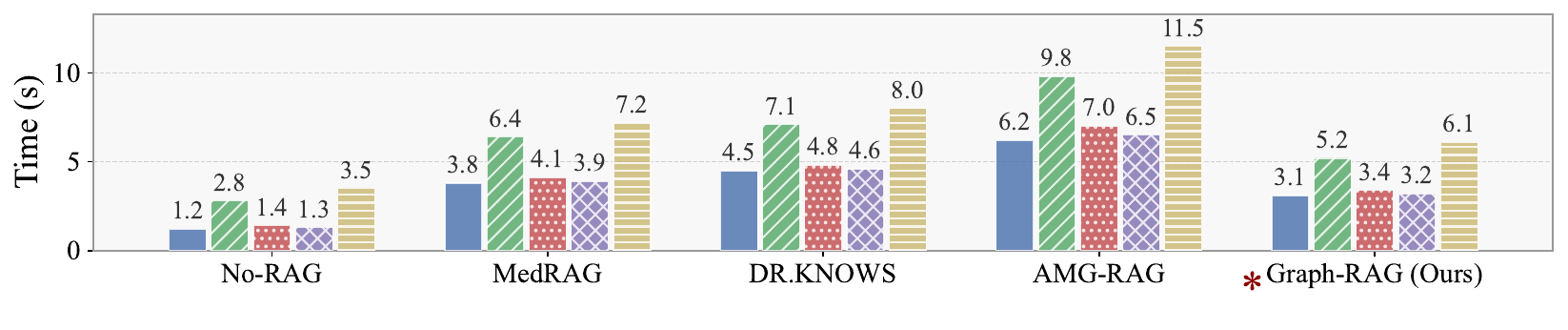}
            \vspace{-0.8cm}
    \caption{
Overview of \textbf{\textsc{EpiBench}} Running Time.
\textcolor[HTML]{4C72B0}{Blue} denotes T1 Knowledge QA,
\textcolor[HTML]{55A868}{green} denotes T2 Report Generation,
\textcolor[HTML]{C44E52}{red} denotes T3 Precision Medicine,
\textcolor[HTML]{8172B2}{purple} denotes T4 Treatment Recommendation,
and \textcolor[HTML]{CCB974}{olive} denotes T5 Deep Research.
The \textcolor[HTML]{8B0000}{red star} highlights Graph-RAG (Ours).
}
\vspace{-0.4cm}

    \label{fig:EpiBench_runningtime}
\end{figure*}

\textbf{Baselines.} We compare against three knowledge-augmented systems:
\textit{MedRAG}~\cite{xiong2024benchmarking} (flat dense
retrieval over PubMed and textbooks),
\textit{DR.KNOWS}~\cite{gao2025drknows} (UMLS-based KG paths,
diagnostic tasks only), and
\textit{AMG-RAG}~\cite{rezaei2025amgrag} (dynamic KG
construction via LLM agents, no multi-task evaluation).
None combines a curated domain-specific KG with multi-task
clinical evaluation. As shown in Figure~\ref{fig:overview_result}, Graph-RAG consistently dominates all baselines across all five task forms, with the largest margin on T3 Precision Medicine where domain-specific multi-hop reasoning is essential. Graph-RAG also maintains competitive inference efficiency relative to all baselines
(Figure~\ref{fig:EpiBench_runningtime}).

\begin{table*}[t]
\centering
\setlength{\tabcolsep}{3pt}
\renewcommand{\arraystretch}{1.2}
\caption{Knowledge \& Clinical Reasoning results.
\textbf{T1a}: MCQ, Acc: Top-1 Accuracy (\%).
\textbf{T1b}: Open-ended QA, LJ: LLM-as-Judge (1--5).
\textbf{T3}: Precision Medicine, GC: Guideline Concordance (\%).
\textbf{T4}: Treatment Recommendation, DFS: Drug Safety Score,
KGEC: KG Evidence Coverage.
R-L: ROUGE-L; BS: BERTScore F1; Reas.: Reasoning Accuracy.
Results are mean $\pm$ std; $\Delta$: avg.\ relative improvement.
Claude~S4 refers to Claude Sonnet~4.}
\label{tab:results_knowledge}
\resizebox{\textwidth}{!}{%
\begin{tabular}{ll cccc c cccc c cc c cccc}
\toprule
& &
\multicolumn{4}{c}{\textbf{T1a: MCQ}} & &
\multicolumn{4}{c}{\textbf{T1b: Open-ended QA}} & &
\multicolumn{2}{c}{\textbf{T3: Prec.\ Med.}} & &
\multicolumn{4}{c}{\textbf{T4: Treatment Rec.}} \\
\cmidrule{3-6}\cmidrule{8-11}\cmidrule{13-14}\cmidrule{16-19}
\textbf{Model} & \textbf{RAG} &
\textbf{Acc} & \textbf{R-L} & \textbf{BS} & \textbf{Reas.} & &
\textbf{LJ} & \textbf{R-L} & \textbf{BS} & \textbf{Reas.} & &
\textbf{Acc} & \textbf{GC} & &
\textbf{Acc} & \textbf{DFS} & \textbf{GC} & \textbf{KGEC} \\
\midrule
\textit{GPT-4o} & --
  & 68.0$\pm$1.4 & 0.38$\pm$0.02 & 0.73$\pm$0.01 & 0.46$\pm$0.03 &
  & 3.61$\pm$0.08 & 0.34$\pm$0.02 & 0.69$\pm$0.01 & 0.35$\pm$0.03 &
  & 53.0$\pm$3.2 & 51.0$\pm$3.5 &
  & 72.0$\pm$2.1 & 0.38$\pm$0.03 & 68.0$\pm$2.4 & 0.73$\pm$0.02 \\
\rowcolor{blue!5}
& w/
  & \textbf{75.0$\pm$1.2} & 0.41$\pm$0.02 & 0.80$\pm$0.01 & 0.59$\pm$0.02 &
  & \textbf{4.33$\pm$0.06} & 0.38$\pm$0.02 & 0.74$\pm$0.01 & 0.40$\pm$0.02 &
  & 69.0$\pm$2.6 & 65.0$\pm$2.8 &
  & \textbf{81.0$\pm$1.7} & 0.41$\pm$0.03 & 76.0$\pm$2.0 & 0.77$\pm$0.02 \\
\textit{Claude S4} & --
  & 69.0$\pm$1.3 & 0.41$\pm$0.02 & 0.77$\pm$0.01 & 0.49$\pm$0.03 &
  & 3.74$\pm$0.07 & 0.39$\pm$0.02 & 0.75$\pm$0.01 & 0.34$\pm$0.03 &
  & 66.0$\pm$2.8 & 61.0$\pm$3.0 &
  & 70.0$\pm$2.2 & 0.41$\pm$0.03 & 66.0$\pm$2.5 & 0.74$\pm$0.02 \\
\rowcolor{blue!5}
& w/
  & 73.0$\pm$1.2 & \textbf{0.45$\pm$0.01} & \textbf{0.82$\pm$0.01} & \textbf{0.64$\pm$0.02} &
  & 4.21$\pm$0.05 & \textbf{0.42$\pm$0.01} & \textbf{0.80$\pm$0.01} & \textbf{0.42$\pm$0.02} &
  & \textbf{82.0$\pm$2.1} & \textbf{79.0$\pm$2.3} &
  & 79.0$\pm$1.8 & \textbf{0.45$\pm$0.02} & \textbf{78.0$\pm$1.9} & \textbf{0.82$\pm$0.01} \\
\textit{Gemini} & --
  & 63.0$\pm$1.5 & 0.36$\pm$0.02 & 0.74$\pm$0.01 & 0.47$\pm$0.03 &
  & 3.54$\pm$0.09 & 0.35$\pm$0.02 & 0.71$\pm$0.02 & 0.37$\pm$0.03 &
  & 58.0$\pm$3.1 & 51.0$\pm$3.4 &
  & 65.0$\pm$2.4 & 0.36$\pm$0.03 & 61.0$\pm$2.7 & 0.67$\pm$0.02 \\
\rowcolor{blue!5}
& w/
  & 68.0$\pm$1.4 & 0.42$\pm$0.02 & 0.81$\pm$0.01 & 0.56$\pm$0.02 &
  & 3.96$\pm$0.07 & 0.37$\pm$0.02 & 0.76$\pm$0.01 & 0.41$\pm$0.02 &
  & 74.0$\pm$2.5 & 74.0$\pm$2.6 &
  & 74.0$\pm$2.0 & 0.42$\pm$0.03 & 72.0$\pm$2.2 & 0.76$\pm$0.02 \\
\textit{Llama} & --
  & 57.0$\pm$1.6 & 0.33$\pm$0.02 & 0.70$\pm$0.02 & 0.30$\pm$0.03 &
  & 3.42$\pm$0.10 & 0.29$\pm$0.02 & 0.63$\pm$0.02 & 0.29$\pm$0.03 &
  & 44.0$\pm$3.5 & 48.0$\pm$3.3 &
  & 58.0$\pm$2.6 & 0.33$\pm$0.04 & 55.0$\pm$2.8 & 0.70$\pm$0.02 \\
\rowcolor{blue!5}
& w/
  & 66.0$\pm$1.4 & 0.36$\pm$0.02 & 0.76$\pm$0.01 & 0.41$\pm$0.03 &
  & 3.95$\pm$0.07 & 0.35$\pm$0.02 & 0.66$\pm$0.01 & 0.34$\pm$0.03 &
  & 60.0$\pm$2.9 & 65.0$\pm$2.8 &
  & 67.0$\pm$2.2 & 0.36$\pm$0.03 & 64.0$\pm$2.4 & 0.74$\pm$0.02 \\
\textit{Qwen} & --
  & 59.0$\pm$1.5 & 0.32$\pm$0.02 & 0.67$\pm$0.02 & 0.28$\pm$0.04 &
  & 3.44$\pm$0.11 & 0.27$\pm$0.03 & 0.61$\pm$0.02 & 0.24$\pm$0.04 &
  & 41.0$\pm$3.6 & 44.0$\pm$3.4 &
  & 55.0$\pm$2.7 & 0.32$\pm$0.04 & 52.0$\pm$2.9 & 0.63$\pm$0.03 \\
\rowcolor{blue!5}
& w/
  & 64.0$\pm$1.4 & 0.37$\pm$0.02 & 0.74$\pm$0.01 & 0.39$\pm$0.03 &
  & 3.74$\pm$0.08 & 0.33$\pm$0.02 & 0.69$\pm$0.01 & 0.31$\pm$0.03 &
  & 58.0$\pm$3.0 & 61.0$\pm$2.9 &
  & 64.0$\pm$2.3 & 0.37$\pm$0.03 & 62.0$\pm$2.5 & 0.69$\pm$0.02 \\
\textit{Mistral} & --
  & 51.0$\pm$1.8 & 0.31$\pm$0.03 & 0.64$\pm$0.02 & 0.24$\pm$0.04 &
  & 3.54$\pm$0.12 & 0.26$\pm$0.03 & 0.58$\pm$0.02 & 0.25$\pm$0.04 &
  & 38.0$\pm$3.8 & 44.0$\pm$3.5 &
  & 51.0$\pm$2.9 & 0.31$\pm$0.04 & 49.0$\pm$3.0 & 0.61$\pm$0.03 \\
\rowcolor{blue!5}
& w/
  & 61.0$\pm$1.5 & 0.35$\pm$0.02 & 0.72$\pm$0.01 & 0.33$\pm$0.03 &
  & 3.85$\pm$0.09 & 0.31$\pm$0.02 & 0.62$\pm$0.02 & 0.30$\pm$0.03 &
  & 51.0$\pm$3.2 & 58.0$\pm$3.1 &
  & 62.0$\pm$2.4 & 0.35$\pm$0.03 & 60.0$\pm$2.6 & 0.68$\pm$0.02 \\
\midrule
\multicolumn{2}{l}{\textbf{Avg.\ $\Delta$}}
  & \textbf{+11.3} & \textbf{+0.04} & \textbf{+0.07} & \textbf{+0.13} &
  & \textbf{+0.51} & \textbf{+0.05} & \textbf{+0.05} & \textbf{+0.06} &
  & \textbf{+34.8} & \textbf{+33.2} &
  & \textbf{+15.6} & \textbf{+12.4} & \textbf{+14.1} & \textbf{+4.2} \\
\bottomrule
\end{tabular}}
\end{table*}
\begin{figure}[t]
    \centering
    \begin{subfigure}[t]{0.245\textwidth}
        \centering
        \includegraphics[width=\linewidth]{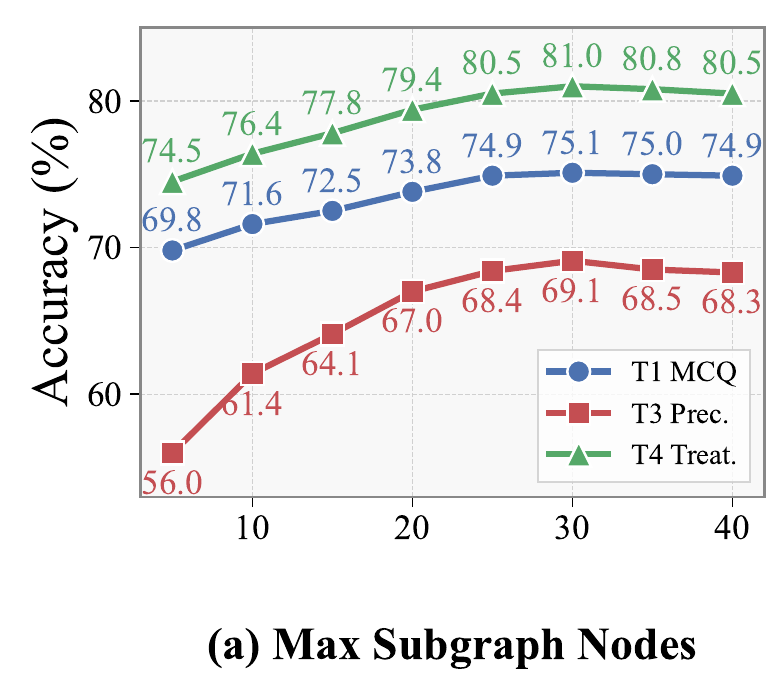}
        \label{fig:sensitivity-a}
    \end{subfigure}
    \hfill
    \begin{subfigure}[t]{0.245\textwidth}
        \centering
        \includegraphics[width=\linewidth]{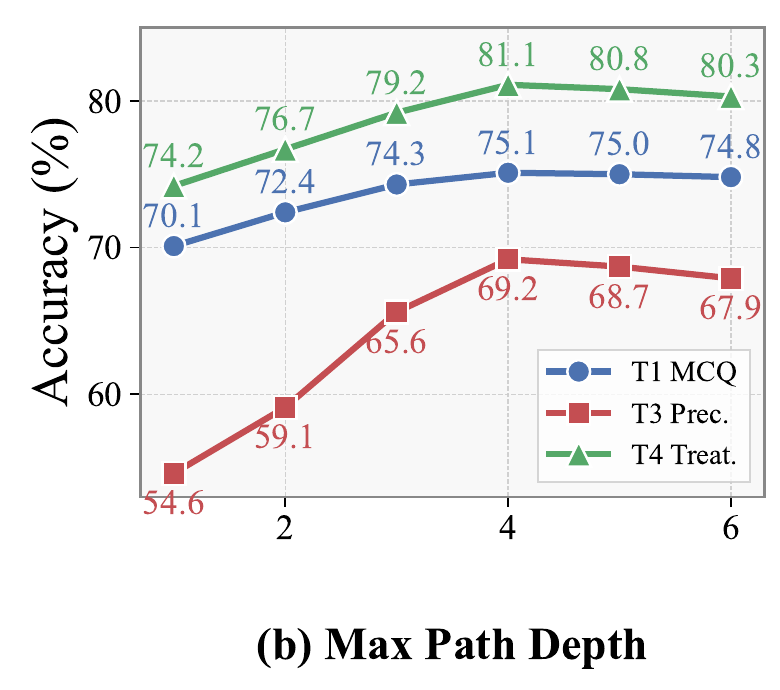}
        \label{fig:sensitivity-b}
    \end{subfigure}
    \hfill
    \begin{subfigure}[t]{0.245\textwidth}
        \centering
        \includegraphics[width=\linewidth]{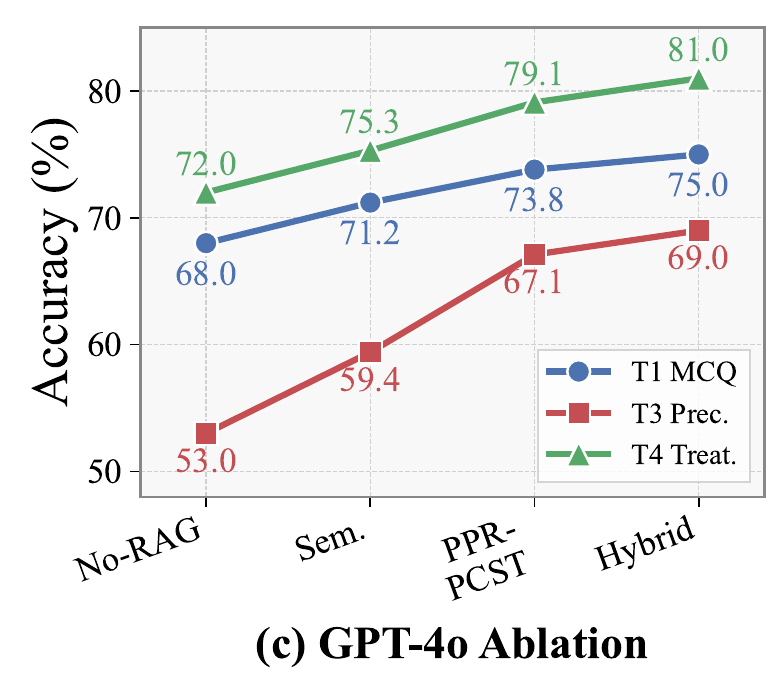}
        \label{fig:sensitivity-c}
    \end{subfigure}
    \hfill
    \begin{subfigure}[t]{0.245\textwidth}
        \centering
        \includegraphics[width=\linewidth]{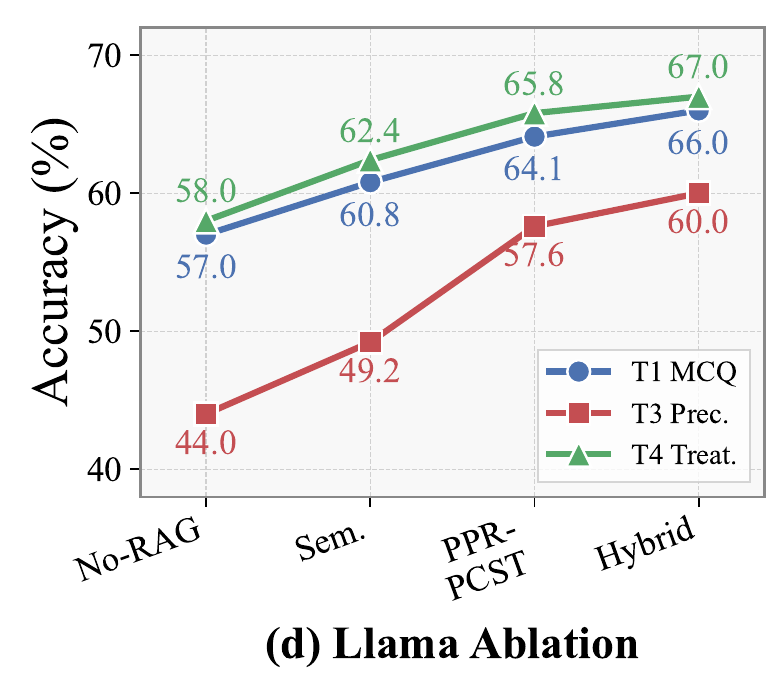}
        \label{fig:sensitivity-d}
    \end{subfigure}

    \vspace{-0.3cm}
    \caption{
    Sensitivity analysis and ablation results of Graph-RAG.
    \textcolor[HTML]{4C72B0}{Blue} denotes T1 MCQ,
    \textcolor[HTML]{C44E52}{red} denotes T3 Precision Medicine,
    and \textcolor[HTML]{55A868}{green} denotes T4 Treatment Recommendation.
    Dashed vertical lines mark the selected optimal hyperparameter settings.
    }
    \label{fig:sensitivity-ablation}
    \vspace{-0.6cm}
\end{figure}

\begin{figure}[t]
    \centering
    \vspace{-3mm}
    \includegraphics[width=\textwidth]{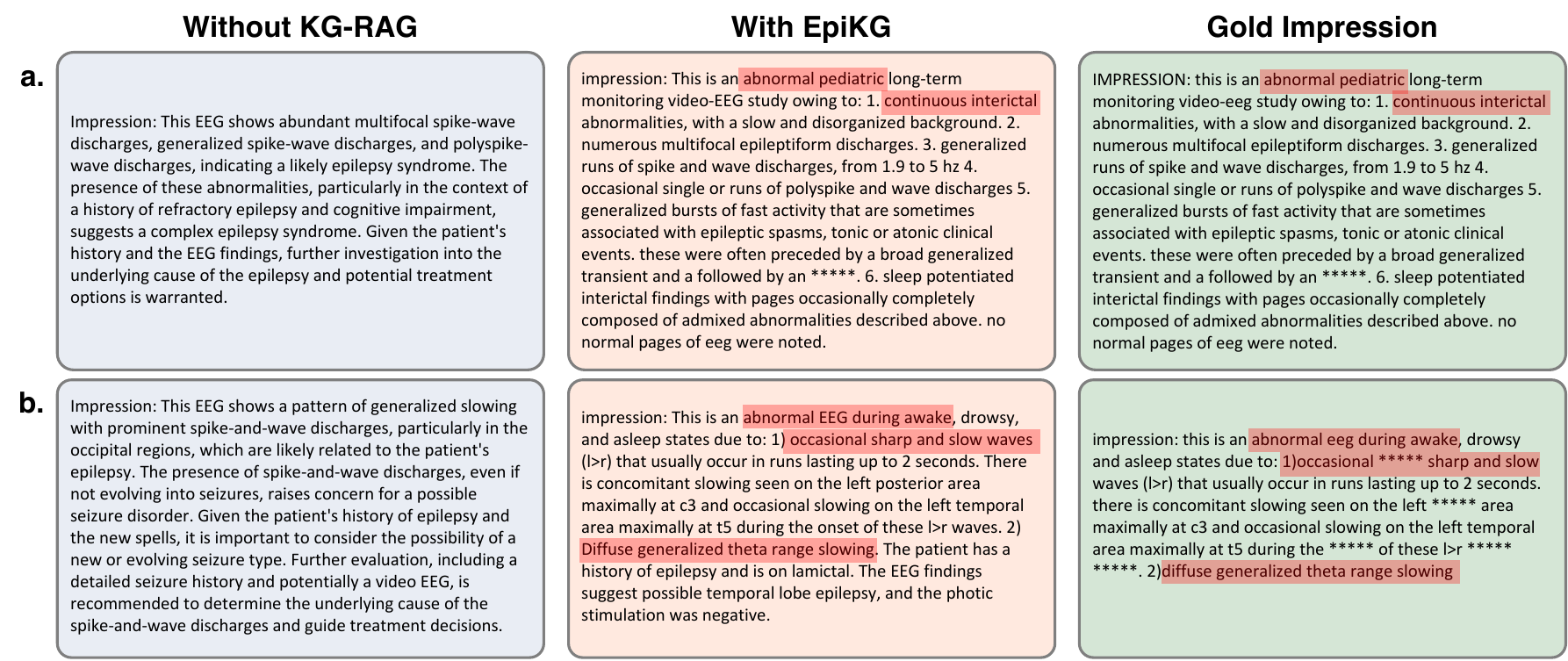}
    \caption{Two impression generation examples on S0001. Three column refer to medgemma model result, medgemma w/ EpiKG, and ground truth report.}
    \label{fig:impressioncase}
    \vspace{-8mm}
\end{figure}

\noindent \textbf{Task 1: Clinical Decision Accuracy (CDA).
(Tables~\ref{tab:results_knowledge})}
Graph-RAG yields consistent MCQ gains across all six
models (avg.\ +11.3 pp). Open-source models benefit
most (Mistral +19.6\%, Llama +15.8\%),
while closed-source gains are smaller (GPT-4o
+10.3\%, Claude S4 +5.8\%), suggesting that
models with stronger parametric epilepsy knowledge
have less marginal benefit from KG augmentation.
Notably, LLM-as-Judge reasoning accuracy improves by +31.9\% on average, substantially larger than ROUGE-L (+14.6\%) and BERTScore (+7.6\%) gains, indicating that Graph-RAG improves the quality of clinical reasoning chains rather than surface-level lexical overlap with reference answers.
\begin{wraptable}{r}{0.43\textwidth}
\vspace{-7pt}
\centering
\caption{\textbf{Generation tasks for neurology reports}
on Harvard Electroencephalography Database
v4.1~\cite{zafar2025harvard,sun2025harvard} processed
using the pipeline
from~\cite{pradeepkumar2026neural}. Results are
reported as mean $\pm$ standard deviation; $\Delta$
denotes relative improvement (\%).}
\label{tab:impression_generation}
\setlength{\tabcolsep}{3pt}
\small
\begin{tabular}{llcc}
\toprule
\textbf{Model} & \textbf{RAG} & \textbf{METEOR} & \textbf{$\Delta$} \\
\midrule
\textit{Gemma-3-4b-it}    & --  & 0.23$\pm$0.06 & -- \\
\rowcolor{blue!5}
                           & w/  & \textbf{0.26$\pm$0.08} & +13.0 \\
\textit{Llama-3.2-3B-it}  & --  & 0.29$\pm$0.08 & -- \\
\rowcolor{blue!5}
                           & w/  & \textbf{0.31$\pm$0.11} & +6.8 \\
\textit{Medgemma-4b-it}   & --  & 0.26$\pm$0.13 & -- \\
\rowcolor{blue!5}
                           & w/  & \textbf{0.34$\pm$0.17} & +30.8 \\
\textit{Qwen3-4B-it-2507} & --  & 0.26$\pm$0.06 & -- \\
\rowcolor{blue!5}
                           & w/  & \textbf{0.27$\pm$0.07} & +3.8 \\
\bottomrule
\end{tabular}
\vspace{-8pt}
\end{wraptable}
On open-ended QA, GPT-4o achieves 4.33/5.0
(+19\% over No-RAG). Graph-RAG outperforms
MedRAG ($\sim$62\%) and AMG-RAG
(66.3\%) by substantial margins, confirming the advantage of domain-specific KG curation over
general-purpose retrieval.

\noindent \textbf{Task 2: Clinical Report Generation (CRG)
(Table~\ref{tab:impression_generation}).}
We evaluate CRG using the Harvard EEG database (site \texttt{s0001}) \cite{zafar2025harvard,sun2025harvard} processed with the CLEM preprocessing pipeline \cite{pradeepkumar2026neural} to extract structured EEG impression data. Each instance consists of EEG descriptions, patient information, and neurologist-written impression reports as ground truth. Due to dataset usage restrictions, evaluation is limited to locally deployed LLMs, including Gemma, Llama, MedGemma, and Qwen.
Table~\ref{tab:impression_generation} shows that generating clinically meaningful EEG impressions remains highly challenging for current LLMs, with overall METEOR scores remaining relatively low. 
Nevertheless, integrating \textsc{EpiKG} consistently improves performance for all models. Specially, MedGemma-4B shows the largest gain (+30.8\%
METEOR), suggesting domain-pretrained models benefit most
when structured clinical context is provided.
Figure~\ref{fig:impressioncase} further presents two representative EEG impression generation examples using MedGemma with and without \textsc{EpiKG}. 
Integrating \textsc{EpiKG} enables the model to produce substantially more precise impressions, including detailed waveform descriptions, abnormal slowing patterns, and clinically relevant interpretations that more closely align with neurologist-written reports.

\noindent \textbf{Task 3: Biomarker-Driven Precision Medicine (BPM)
(Table~\ref{tab:results_knowledge}).}
Graph-RAG produces the largest relative gains across
all tasks. \textit{Claude Sonnet 4} reaches 82\%
(+24\%); open-source models improve even more
dramatically, \textit{Qwen} +42\%, \textit{Llama}
+36\%, reversing the usual closed/open-source
performance gap and suggesting that pharmacogenomic
knowledge is more uniformly absent from all LLM
parameter spaces, making KG augmentation equally
essential regardless of model scale.
Critically, without KG context \textit{Mistral}
scores only 38\%, near the 25\% random baseline
for four-option MCQ, confirming that pharmacogenomic
reasoning is not encoded in general-purpose LLM
parameters and cannot be elicited through prompting
alone.

\noindent\textbf{Task 4: Treatment Recommendation (TR)
(Table~\ref{tab:results_knowledge}).}
On MedQA-USMLE, Graph-RAG improves Top-1 Accuracy
+15.6\%, Drug Safety Score +12.4\%, and Guideline
Concordance +14.1\%. On MMLU Professional Medicine,
DFS and GC improve far more substantially (+28.1\%
and +28.4\%), while raw accuracy gains are similar
(+17.2\%). This divergence suggests that MMLU
questions, which require broader multi-step clinical
reasoning, expose safety and guideline gaps that
narrow factual recall tasks do not, and that
Graph-RAG's primary value in treatment recommendation
\begin{wraptable}{r}{0.46\textwidth}
\vspace{-8pt}
\centering
\caption{\textbf{Deep Epi-Research results.}
LJ: LLM-as-Judge (1--5); R-L: ROUGE-L.
Results are reported as mean $\pm$ standard
deviation; $\Delta$ denotes relative
improvement (\%).}
\label{tab:rq5}
\setlength{\tabcolsep}{3pt}
\scriptsize
\begin{tabular}{llcccc}
\toprule
\textbf{Model} & \textbf{RAG} &
\textbf{LJ} & \textbf{$\Delta$} &
\textbf{R-L} & \textbf{$\Delta$} \\
\midrule
\textit{GPT-4o}    & --  & 3.56$\pm$0.12 & --    & 0.34$\pm$0.02 & -- \\
\rowcolor{blue!5}
                   & w/  & \textbf{4.25$\pm$0.09} & +19.4 & 0.41$\pm$0.02 & +20.6 \\
\textit{Claude S4} & --  & 3.69$\pm$0.11 & --    & 0.36$\pm$0.02 & -- \\
\rowcolor{blue!5}
                   & w/  & 4.13$\pm$0.08 & +11.9 & \textbf{0.43$\pm$0.02} & +19.4 \\
\textit{Gemini}    & --  & 3.49$\pm$0.13 & --    & 0.33$\pm$0.03 & -- \\
\rowcolor{blue!5}
                   & w/  & 3.88$\pm$0.10 & +11.2 & 0.39$\pm$0.02 & +18.2 \\
\textit{Llama}     & --  & 3.37$\pm$0.14 & --    & 0.31$\pm$0.03 & -- \\
\rowcolor{blue!5}
                   & w/  & 3.87$\pm$0.11 & +14.8 & 0.37$\pm$0.02 & +19.4 \\
\textit{Qwen}      & --  & 3.39$\pm$0.15 & --    & 0.32$\pm$0.03 & -- \\
\rowcolor{blue!5}
                   & w/  & 3.66$\pm$0.12 & +8.0  & 0.36$\pm$0.02 & +12.5 \\
\textit{Mistral}   & --  & 3.49$\pm$0.16 & --    & 0.30$\pm$0.03 & -- \\
\rowcolor{blue!5}
                   & w/  & 3.77$\pm$0.13 & +8.0  & 0.35$\pm$0.03 & +16.7 \\
\bottomrule
\end{tabular}
\vspace{-8pt}
\end{wraptable}
is improving clinical safety rather than answer
correctness per se. The positive correlation between
KGEC and DFS improvements (+4.2\% vs +13.2\% across
the two datasets) further confirms that stronger KG
utilisation directly translates to better
contraindication avoidance.

\noindent\textbf{Task 5: Deep Research Planning (DRP)
(Table~\ref{tab:rq5}).}
Graph-RAG improves LLM-as-Judge scores by +12.2\%
on average; \textit{GPT-4o} achieves 4.25/5.0
(+19.4\%). The closed/open-source performance gap
narrows substantially relative to T1 and T3:
\textit{Llama} reaches 3.87 vs \textit{GPT-4o}'s
4.25 under Graph-RAG (gap of 0.38), compared to a
gap of 9 pp on T1 MCQ. This convergence suggests
that research plan generation depends more on
structural reasoning ability, the capacity to
follow retrieved KG paths and synthesise coherent
hypotheses, than on parametric domain knowledge,
and that Graph-RAG can substantially fill the
capability gap between open and closed source
models on this dimension.

\begin{figure*}[t]
    \centering
    \vspace{-0.2cm}
\includegraphics[width=\linewidth]{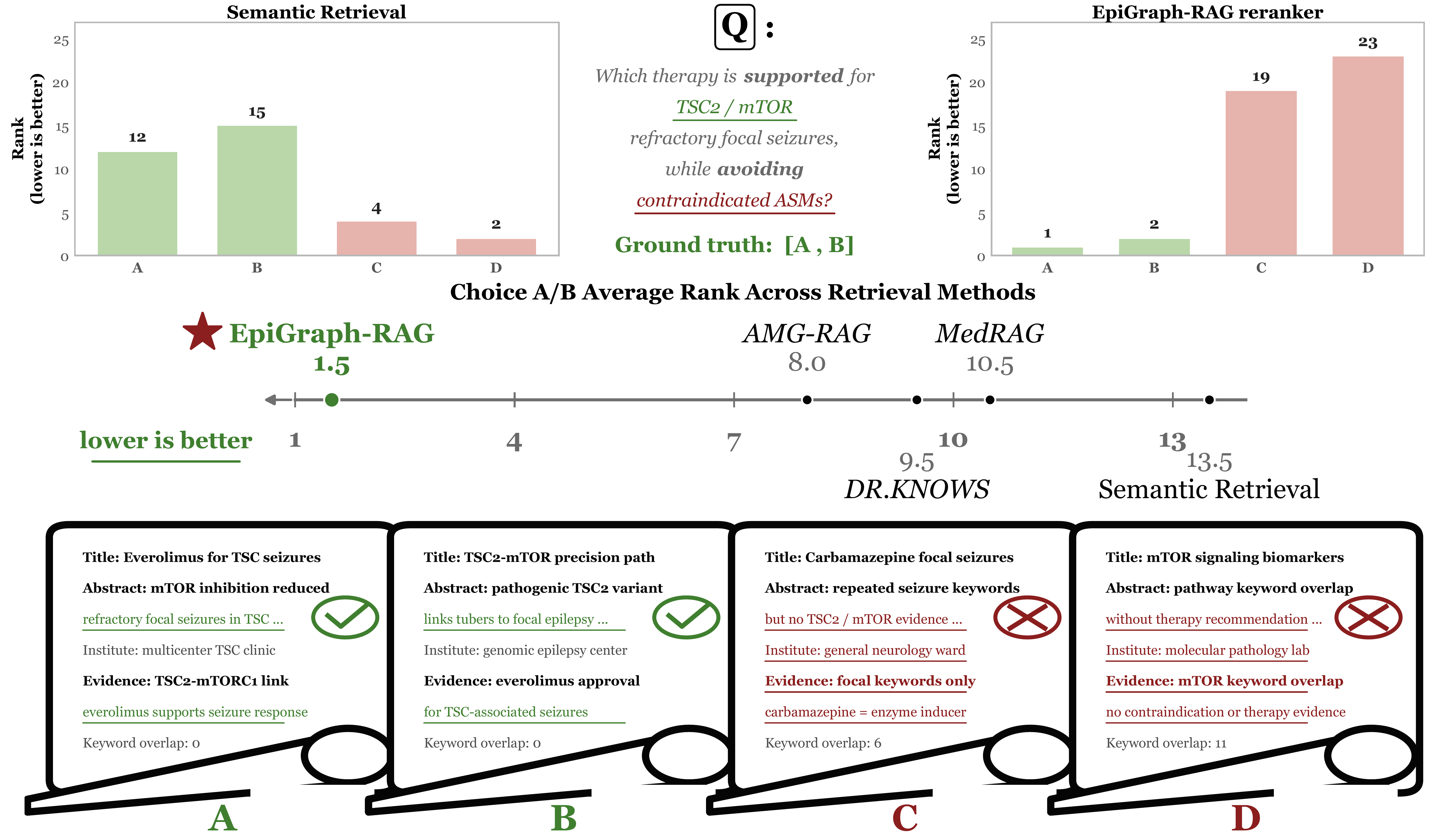}
    \caption{
Case study of a TSC2 precision-medicine query.
Semantic Retrieval ranks distractor candidates \textcolor{darkred}{C} and \textcolor{darkred}{D} highly due to surface keyword overlap, whereas EpiGraph-RAG reranks the true supporting evidence \textcolor{gtgreen}{A} and \textcolor{gtgreen}{B} to the top by following the \textcolor{gtgreen}{TSC2/mTOR} therapeutic path and avoiding \textcolor{darkred}{contraindicated ASMs}.
\textcolor{gtgreen}{Green} denotes ground-truth supporting evidence, and \textcolor{darkred}{dark red} denotes non-ground-truth evidence.
}
    \label{fig:caseshow} 
    \vspace{-0.2cm}
\end{figure*}

\noindent\textbf{Ablation and Sensitivity
(Figure~\ref{fig:sensitivity-ablation}).}
PPR-PCST outperforms semantic retrieval by
+2.6--7.7 pp on T1 and +7.7--8.4 pp on T3;
Hybrid adds a further +1.2--2.4 pp. The advantage
of graph-based retrieval is most pronounced on T3,
where multi-hop reasoning from gene to drug
mechanism requires traversing KG topology that
flat retrieval cannot exploit. Optimal subgraph
size is 30 nodes and path depth is 4 hops; T3
shows the strongest depth sensitivity (depth 2:
58.7\% vs depth 4: 69.0\% for GPT-4o), confirming
that pharmacogenomic reasoning inherently requires
3--4 hop chains.

\section{Conclusion and Discussion}
\label{sec:discussion}

\textsc{EpiKG} and \textsc{EpiBench} together establish
a knowledge-grounded evaluation framework \textbf{\textsc{EpiGraph}} for
epilepsy. \textsc{EpiKG} provides the first
open-source domain-specific epilepsy KG integrating
seven ontologies across five clinical layers, and
\textsc{EpiBench} demonstrates that Graph-RAG over
\textsc{EpiKG} consistently improves LLM performance
across all five clinical task forms relative to both
no-retrieval baselines and general-purpose retrieval
systems. \textsc{EpiGraph} raises a number of unique and novel problems based on the difficulty of its tasks:  How can pharmacogenomic knowledge be encoded in LLM parameters rather than retrieved at inference time? How can generated clinical text be aligned to neurologist-level expression norms beyond automated metrics? Is Graph-RAG primarily supplying missing domain knowledge or providing reasoning structure that weaker models lack?
We hope that given the scale of \textsc{EpiKG} and the
task diversity of \textsc{EpiBench}, it can be used to
explore retrieval-augmented fine-tuning or memory
approaches~\cite{lewis2020retrieval,zhang2022greaselm} that can encode
epilepsy-specific clinical rules from large numbers of
KG triplets and literature entries.
Further, we hope the inclusion of neurologist gold
standards in T2 and expert research plan annotations
in T5 will spur work into evaluating clinical language
generation against human-level judgements rather than
solely on automated proxy metrics.
\textsc{EpiGraph} set a new standard
for clinical AI evaluation and open new challenges for modern LLMs to drive future development at the intersection of clinical neurology and biomedical reasoning.

\bibliographystyle{unsrt}   
\bibliography{references}   

\newpage
\appendix
\appendix

\section*{Appendix Contents}

\begin{center}
\begin{tabular}{@{}p{0.82\textwidth}r@{}}
\textbf{A\quad Related Work} & \pageref{sec:app_related} \\[4pt]
\textbf{B\quad EpiKG Construction Details} & \pageref{sec:app_epikg} \\
\quad B.1\enspace Ontology Sources and Coverage & \pageref{sec:app_epikg_ontology} \\
\quad B.2\enspace Entity Normalisation Protocol & \pageref{sec:app_epikg_norm} \\
\quad B.3\enspace Relation Extraction Details & \pageref{sec:app_epikg_relation} \\
\quad B.4\enspace Knowledge Graph Statistics & \pageref{sec:app_epikg_stats} \\[4pt]
\textbf{C\quad EpiBench Dataset Details} & \pageref{sec:app_epibench} \\
\quad C.1\enspace Dataset Construction Protocols & \pageref{sec:app_epibench_data} \\
\quad C.2\enspace Gold Standard Annotation & \pageref{sec:app_epibench_gold} \\
\quad C.3\enspace Dataset Statistics & \pageref{sec:app_epibench_stats} \\[4pt]
\textbf{D\quad Graph-RAG Retriever Details} & \pageref{sec:app_retriever} \\
\quad D.1\enspace PPR-PCST Implementation & \pageref{sec:app_retriever_ppr} \\
\quad D.2\enspace Semantic Retrieval Implementation & \pageref{sec:app_retriever_sem} \\
\quad D.3\enspace Hyperparameter Settings & \pageref{sec:app_retriever_hyper} \\[4pt]
\textbf{E\quad Evaluation Metrics} & \pageref{sec:app_metrics} \\[4pt]
\textbf{F\quad Prompts} & \pageref{sec:app_prompts} \\[4pt]
\textbf{G\quad Additional Experimental Results} & \pageref{sec:app_results} \\[4pt]
\textbf{H\quad Limitations and Future Work} & \pageref{sec:app_limitations} \\
\end{tabular}
\end{center}

\clearpage

\section{Related Work}
\label{sec:app_related}

\noindent\textbf{Epilepsy knowledge resources.}
Epilepsy-specific knowledge has been encoded in
various forms, ranging from clinical ontologies to
curated databases.
The ILAE 2022 classification
system~\cite{fisher2014ilae} provides a standardised
taxonomy of seizure types and epilepsy syndromes.
OMIM~\cite{hamosh2005omim} and
HPO~\cite{kohler2021human} encode gene--disease
associations and phenotypic descriptions
respectively, while ChEBI~\cite{hastings2016chebi}
provides chemical identifiers for antiseizure
medications.
Epilepsy-specific ontologies have also been
developed to support semantic interoperability and
clinical text
mining~\cite{sahoo2014epilepsy, sargsyan2023epilepsy}.
CPIC guidelines formalise pharmacogenomic rules
for drug selection based on genetic variants
(\textit{Clinical Pharmacogenomics Implementation
Consortium, cpicpgx.org}).
\textsc{EpiKG} integrates these resources into a
unified relational structure, enabling multi-hop
clinical reasoning that individual ontologies
cannot support in isolation.

\noindent\textbf{Biomedical knowledge graphs.}
Knowledge graphs have been widely adopted in
biomedicine for drug discovery, disease
characterisation, and clinical decision
support~\cite{cui2025review, bonner2022review}.
General-purpose biomedical KGs such as
UMLS~\cite{bodenreider2004umls} and
PrimeKG~\cite{chandak2023building} provide broad
coverage but lack epilepsy-specific relation types
and clinical granularity.
Domain-specific KGs have been applied across
biomedical
domains~\cite{lu2025biomedical},
but no comparable resource exists for epilepsy.
\textsc{EpiKG} fills this gap by combining ontology
seeding with LLM-based relation extraction to
produce a clinically grounded, epilepsy-specific KG.

\noindent\textbf{Graph-RAG and knowledge-augmented LLMs.}
Retrieval-augmented generation has emerged as a
dominant paradigm for grounding LLM outputs in
external knowledge~\cite{lewis2020retrieval}.
Graph-based retrieval extends flat document
retrieval by exploiting relational structure to
recover multi-hop reasoning
chains~\cite{yasunaga2021qa, zhang2022greaselm}.
Subgraph-based context augmentation has been shown
to improve factual grounding and reasoning
coherence over dense retrieval
alone~\cite{sun2023think, sen2023knowledge}.
Medical applications include
DR.KNOWS~\cite{gao2025drknows},
MedRAG~\cite{xiong2024benchmarking}, and
AMG-RAG~\cite{rezaei2025amgrag}, which apply
KG-augmented retrieval to general clinical tasks.
More recently, MedReason~\cite{chen2025medreason}
demonstrates that KG-elicited reasoning steps
improve factual accuracy in medical LLMs.
\textsc{EpiGraph} differs from these systems by
combining a curated domain-specific KG with a
multi-task evaluation benchmark, enabling
controlled ablation of retrieval quality across
diverse epilepsy clinical reasoning tasks.

\noindent\textbf{Clinical NLP benchmarks.}
Existing biomedical NLP benchmarks include
MedQA~\cite{jin2021disease},
MMLU Professional Medicine~\cite{hendrycks2020measuring},
PubMedQA~\cite{jin2019pubmedqa}, and
MedMCQA~\cite{pal2022medmcqa}, which evaluate
factual recall and clinical reasoning on general
medical knowledge. Epilepsy-specific evaluation has
been limited to narrow tasks such as seizure
classification from EEG signals~\cite{tveit2023automated}
or pharmacogenomic variant
interpretation~\cite{kuo2024pharmacogenetics}.
\textsc{EpiBench} is the first benchmark to provide
multi-task evaluation of LLMs across the full
epilepsy clinical reasoning pipeline, from syndrome
identification and treatment recommendation to
research planning.

\noindent\textbf{LLM evaluation in clinical settings.}
Recent work has evaluated LLMs on clinical
tasks~\cite{singhal2023large, singhal2025toward},
finding that even large models struggle with
specialised clinical knowledge and guideline
adherence. LLM-as-Judge approaches have been
proposed for evaluating open-ended clinical
generation~\cite{chen2025benchmarking}, and
domain-specific metrics such as guideline
concordance have been introduced to capture
clinically meaningful performance beyond standard
NLP
metrics~\cite{yu2024evaluation}. \textsc{EpiBench}
adopts and extends these evaluation practices with
epilepsy-specific metrics including Drug Safety
Score and KG Evidence Coverage.

\clearpage

\section{EpiKG Construction Details}
\label{sec:app_epikg}

\subsection{Ontology Sources and Coverage}
\label{sec:app_epikg_ontology}

Table~\ref{tab:app_ontology_sources} lists the seven
ontology sources integrated in \textsc{EpiKG},
together with the entity layer each source
contributes, the number of seed entities extracted,
and the license under which each resource is used.

\begin{table}[h]
\centering
\small
\setlength{\tabcolsep}{5pt}
\renewcommand{\arraystretch}{1.2}
\caption{Ontology sources used in \textsc{EpiKG}
construction. Seed counts are after epilepsy-specific
filtering.}
\label{tab:app_ontology_sources}
\begin{tabular}{llcl}
\toprule
\textbf{Source} & \textbf{Layer} &
\textbf{Seed entities} & \textbf{License} \\
\midrule
ILAE 2022  & L1 Syndrome   & 312   & Open access \\
MeSH       & L1, L2, L5    & 2,841 & Public domain \\
HPO        & L2, L5        & 1,654 & CC-BY 4.0 \\
OMIM       & L3 Gene       & 3,892 & Academic use \\
HGNC       & L3 Gene       & 4,107 & CC-BY 4.0 \\
ChEBI      & L4 Treatment  & 2,876 & CC-BY 4.0 \\
AES 2024   & L4 Treatment  & 643   & Open access \\
\midrule
\textbf{Total} & & \textbf{24,324} & \\
\bottomrule
\end{tabular}
\end{table}

\subsection{Entity Normalisation Protocol}
\label{sec:app_epikg_norm}

Entity normalisation maps surface mentions extracted
from literature to canonical ontology identifiers.
The pipeline proceeds in three steps.
First, named entity recognition using a biomedical
NER model identifies entity spans in epilepsy
full-text articles.
Second, candidate spans are matched to ontology
entries via exact string matching, then fuzzy
matching with a threshold of 0.85, and finally
UMLS CUI lookup for unresolved spans.
Third, aliases and abbreviations are resolved
through curated synonym lists derived from each
source ontology.
Entities that cannot be normalised to a canonical
identifier after all three steps are discarded.

\subsection{Relation Extraction Details}
\label{sec:app_epikg_relation}

\noindent\textbf{Rule-based extraction.}
Pattern templates are defined for each of the six
relation types. Each template specifies a syntactic
pattern (subject entity layer, trigger phrase set,
object entity layer) applied to dependency-parsed
sentences containing co-occurring entity pairs.
Table~\ref{tab:app_relation_patterns} lists the
trigger phrase sets for each relation type.

\begin{table}[h]
\centering
\small
\setlength{\tabcolsep}{5pt}
\renewcommand{\arraystretch}{1.2}
\caption{Rule-based extraction trigger phrases
for each relation type.}
\label{tab:app_relation_patterns}
\begin{tabular}{lll}
\toprule
\textbf{Relation} & \textbf{Subject layer} &
\textbf{Trigger phrases (examples)} \\
\midrule
treats               & Treatment & recommended for, first-line, effective in \\
contraindicated\_with & Treatment & avoid, contraindicated, not recommended \\
associated\_with     & Gene      & associated with, linked to, implicated in \\
characteristic\_of   & Diagnostic& characteristic of, consistent with, seen in \\
encodes              & Gene      & encodes, produces, results in \\
expressed\_in        & Gene      & expressed in, detected in, localised to \\
\bottomrule
\end{tabular}
\end{table}

\noindent\textbf{LLM-based extraction.}
MiniMax-Text-01~\cite{minimax2025minimax01} is
prompted with a structured template specifying the
six relation types, their definitions, and the
five entity layers. The prompt instructs the model
to extract triplets in the form
\texttt{(head entity, relation, tail entity)} from
each full-text passage, restricted to entity pairs
where both head and tail have been normalised to
canonical identifiers. Extracted triplets are
deduplicated and merged with rule-based extractions;
conflicts are resolved by retaining the triplet with
the higher paper count $\mathcal{P}$.

\subsection{Knowledge Graph Statistics}
\label{sec:app_epikg_stats}

\begin{table}[h]
\centering
\small
\setlength{\tabcolsep}{5pt}
\renewcommand{\arraystretch}{1.2}
\caption{\textsc{EpiKG} statistics by entity layer
and relation type.}
\label{tab:app_epikg_stats}
\begin{tabular}{lcc}
\toprule
\textbf{Category} & \textbf{Count} &
\textbf{Avg.\ paper count $\mathcal{P}$} \\
\midrule
\multicolumn{3}{l}{\textit{Entity layers}} \\
L1 Syndrome   & 312    & 6.3 \\
L2 Diagnostic & 1,654  & 4.8 \\
L3 Gene       & 4,107  & 5.2 \\
L4 Treatment  & 3,412  & 7.1 \\
L5 Outcome    & 2,198  & 3.9 \\
Protein       & 5,841  & 4.1 \\
Anatomy       & 4,683  & 3.6 \\
Other         & 2,117  & 2.8 \\
\textbf{Total nodes} & \textbf{24,324} & \textbf{4.6} \\
\midrule
\multicolumn{3}{l}{\textit{Relation types}} \\
treats                & 8,214  & 4.2 \\
contraindicated\_with & 1,843  & 3.1 \\
associated\_with      & 12,307 & 3.8 \\
characteristic\_of    & 3,891  & 2.9 \\
encodes               & 4,012  & 2.6 \\
expressed\_in         & 1,742  & 2.4 \\
\textbf{Total edges}  & \textbf{32,009} & \textbf{3.0 (median)} \\
\bottomrule
\end{tabular}
\end{table}

\clearpage

\section{EpiBench Dataset Details}
\label{sec:app_epibench}

\subsection{Dataset Construction Protocols}
\label{sec:app_epibench_data}

\noindent\textbf{T1 EpiBench-MCQ and EpiBench-QA.}
Questions are generated from 2025--2026 epilepsy
papers retrieved from PubMed using epilepsy-specific
MeSH terms, restricted to papers published after the
\textsc{EpiKG} construction cutoff to prevent
knowledge leakage. MCQ distractors are generated by
substituting semantically similar but clinically
incorrect entities from \textsc{EpiKG}. Open-ended
questions are generated by prompting GPT-4.1-mini
with the paper abstract and instructing it to
produce factual questions whose answers are
explicitly stated in the source text.

\noindent\textbf{T2 Harvard EEG dataset.}
EEG text descriptions and computed statistics
(band power, spike rate) are extracted from the
Harvard Electroencephalography Database
v4.1~\cite{zafar2025harvard, sun2025harvard} using
the preprocessing pipeline
from~\cite{pradeepkumar2026neural}. Neurologist-written
clinical impressions are used as gold standards
without modification.

\noindent\textbf{T3 Pharmacogenomic MCQs.}
151 MCQs are constructed by clinical experts from
established gene--drug rules in CPIC guidelines and
ILAE 2022 gene-specific recommendations, spanning
six pharmacogenomic categories: ion channel, mTOR
pathway, metabolic, pharmacokinetic safety,
EEG-guided, and multi-hop reasoning. Each question
requires selecting one ASM from four candidates;
distractors are drawn from the same drug class to
require mechanistic discrimination.

\noindent\textbf{T4 MedQA-USMLE and MMLU.}
Epilepsy-relevant questions are filtered from
MedQA-USMLE~\cite{jin2021disease} and MMLU
Professional Medicine~\cite{hendrycks2020measuring}
using keyword matching on epilepsy syndromes, ASM
names, and EEG findings. Questions not resolvable
against ILAE 2022 or CPIC guidelines are excluded.

\noindent\textbf{T5 PMC Research Planning.}
163 epilepsy full-text papers are sampled from
PubMed Central using epilepsy MeSH terms, restricted
to original research articles published between
2020--2024. Expert annotations are provided by
domain collaborators for 30 papers; LLM-as-Judge
evaluations cover the remaining 133.

\subsection{Gold Standard Annotation}
\label{sec:app_epibench_gold}

Expert annotations for T5 are produced by
neurologists with $>$5 years of epilepsy research
experience. Annotators are provided with the full
paper text and instructed to write (i) a focused
research question the paper addresses, (ii) a
study design rationale, and (iii) a list of required
data sources. Inter-annotator agreement is measured
on a 20-paper overlap subset using Cohen's $\kappa$;
$\kappa = 0.81$ for research question
quality and $\kappa = 0.76$ for
feasibility ratings.

\subsection{Dataset Statistics}
\label{sec:app_epibench_stats}

\begin{table}[h]
\centering
\small
\setlength{\tabcolsep}{5pt}
\renewcommand{\arraystretch}{1.2}
\caption{\textsc{EpiBench} dataset statistics
by task.}
\label{tab:app_epibench_stats}
\begin{tabular}{llccc}
\toprule
\textbf{Task} & \textbf{Dataset} &
\textbf{Instances} & \textbf{Gold standard} &
\textbf{Split} \\
\midrule
T1a MCQ    & EpiBench-MCQ       & 1,000  & Source papers              & Test only \\
T1b QA     & EpiBench-QA        & 5,199  & Reference answers          & Test only \\
T2 Report  & Harvard EEG v4.1   & 40,000+& Neurologist impressions    & Test only \\
T3 Prec.   & CPIC/ILAE MCQ      & 151    & Guidelines                 & Test only \\
T4 Treat.  & MedQA-USMLE        & 200 & Guidelines              & Test only \\
T4 Treat.  & MMLU Prof.\ Med.   & 272    & Guidelines                 & Test only \\
T5 Research & PMC papers        & 163    & Expert (30) + LLM-Judge (133) & Test only \\
\bottomrule
\end{tabular}
\end{table}

\clearpage

\section{Graph-RAG Retriever Details}
\label{sec:app_retriever}

\subsection{PPR-PCST Implementation}
\label{sec:app_retriever_ppr}

The PPR-PCST retriever proceeds in three stages.
First, named entity recognition identifies seed
entities $\mathcal{S}$ in the query using the same
biomedical NER model used in \textsc{EpiKG}
construction.
Second, Personalized PageRank~\cite{yasunaga2021qa}
is run from $\mathcal{S}$ over the \textsc{EpiKG}
adjacency matrix with restart probability
$\alpha = 0.15$ and damping factor $1 - \alpha$,
producing a relevance score $r(v)$ for each node
$v$.
Third, a Prize-Collecting Steiner Tree
(PCST)~\cite{zhang2022greaselm} approximation
extracts a connected subgraph by assigning prize
$r(v)$ to each node and minimising edge costs,
subject to a maximum node budget of 30 and maximum
path depth of 4 hops.
The extracted subgraph is serialised into structured
reasoning paths by enumerating all source-to-sink
paths and formatting each as
\texttt{(head, relation[Np], tail)} where $N$ is
the paper count annotation.

\subsection{Semantic Retrieval Implementation}
\label{sec:app_retriever_sem}

The semantic retriever encodes the query using
\texttt{all-MiniLM-L6-v2}~\cite{SBERT} and retrieves
the top-$k$ most similar \textsc{EpiKG} nodes by
cosine similarity over pre-computed node embeddings.
Node embeddings are constructed by encoding the
concatenation of the entity name and its ontology
definition. Local neighbourhoods (depth 1) of the
top-$k$ nodes are extracted and serialised in the
same format as PPR-PCST paths. $k = 10$ is used
in all experiments.

\subsection{Hyperparameter Settings}
\label{sec:app_retriever_hyper}

\begin{table}[h]
\centering
\small
\setlength{\tabcolsep}{5pt}
\renewcommand{\arraystretch}{1.2}
\caption{Graph-RAG retriever hyperparameters.
Optimal values are selected based on T1 MCQ
validation performance.}
\label{tab:app_retriever_hyper}
\begin{tabular}{lcc}
\toprule
\textbf{Hyperparameter} & \textbf{Range tested} &
\textbf{Selected value} \\
\midrule
Max subgraph nodes  & 10, 20, 30, 40 & 30 \\
Max path depth      & 2, 3, 4, 5     & 4  \\
PPR restart prob.\ $\alpha$ & 0.10, 0.15, 0.20 & 0.15 \\
Semantic top-$k$    & 5, 10, 15      & 10 \\
NER confidence threshold & 0.7, 0.8, 0.9 & 0.8 \\
\bottomrule
\end{tabular}
\end{table}

\clearpage

\section{Evaluation Metrics}
\label{sec:app_metrics}

\noindent\textbf{Top-1 Accuracy.}
For MCQ tasks, the model prediction is the option
letter with the highest generation probability or
the option explicitly named in the generated
response. Accuracy is the proportion of correct
predictions over all instances.

\noindent\textbf{ROUGE-L.}
Longest common subsequence F1 between the generated
output and the reference answer~\cite{lin2004rouge}.

\noindent\textbf{BERTScore F1.}
Token-level semantic similarity between generated
and reference text using contextual
embeddings~\cite{zhang2019bertscore}.
We use \texttt{roberta-large} as the backbone model.

\noindent\textbf{LLM-as-Judge.}
GPT-4.1-mini scores generated outputs on a 1--5
Likert scale across three dimensions: factual
correctness, clinical relevance, and reasoning
quality~\cite{chen2025benchmarking}. The final
score is the average across dimensions.

\noindent\textbf{Clinical NER F1.}
Entity-level F1 measuring overlap between syndrome,
drug, and finding mentions in generated outputs and
reference texts, using the biomedical NER model
from \textsc{EpiKG} construction.

\noindent\textbf{Hallucination Rate.}
Proportion of generated sentences classified as
contradicting the source document by a Natural
Language Inference (NLI) model fine-tuned on
biomedical text.

\noindent\textbf{Guideline Concordance (GC).}
Binary indicator of whether the recommended
treatment is consistent with ILAE 2022 and CPIC
guidelines for the given syndrome and genetic
context, averaged over all instances.

\noindent\textbf{Drug Safety Score (DFS).}
Proportion of generated treatment recommendations
that do not include any contraindicated ASM for
the given patient context, as defined by CPIC
and ILAE 2022 guidelines.

\noindent\textbf{KG Evidence Coverage (KGEC).}
Proportion of \textsc{EpiKG} entities in the
retrieved subgraph that appear in the generated
output, measuring active utilisation of retrieved
KG context~\cite{yu2024evaluation}.

\noindent\textbf{Alignment Score.}
Composite score measuring consistency between
generated research plans and source papers,
combining LLM-as-Judge ratings on clinical
impression alignment, research question quality,
and feasibility.

\clearpage

\section{EpiBench and EpiKG Documentation}
\label{sec:app_documentation}

\subsection{Dataset Documentation and Intended Use}
\label{sec:app_doc_intended}

\textsc{EpiKG} and \textsc{EpiBench} are intended
for academic research on epilepsy clinical AI,
knowledge graph construction, and LLM evaluation.
Both resources are derived from publicly available
ontologies, clinical guidelines, and open-access
literature; no private patient data are included.
\textsc{EpiKG} and \textsc{EpiBench} are released
with code and documentation on Hugging Face at
\href{https://huggingface.co/RAI-Lab/EpiGraph}{\texttt{https://huggingface.co/RAI-Lab/EpiGraph}}
and on GitHub at
\href{https://github.com/LabRAI/EpiGraph}{\texttt{https://github.com/LabRAI/EpiGraph}}.
The released package includes dataset summary,
data preview, and evaluation scripts.

\noindent\textbf{Data organisation.}
The release is structured as follows:
\begin{itemize}
    \item \texttt{epikg/nodes/}: entity files per
    layer (L1--L5) in JSON format, each record
    containing entity name, canonical identifier,
    ontology source, and layer label.
    \item \texttt{epikg/edges/}: triplet files per
    relation type in JSON format, each record
    containing head entity, relation, tail entity,
    and paper count $\mathcal{P}$.
    \item \texttt{epikg/graph.pkl}: full
    \textsc{EpiKG} graph serialised as a
    NetworkX DiGraph with node and edge attributes.
    \item \texttt{epibench/t1/}: EpiBench-MCQ
    (1,000 questions) and EpiBench-QA (5,199
    questions) in JSON format.
    \item \texttt{epibench/t2/}: a private-data
    adapter for authorized Harvard EEG exports,
    including EEG descriptions, computed statistics
    (band power, spike rate), and neurologist-written
    clinical impressions as gold standards. The
    public release includes schema code rather than
    redistributing restricted patient data.
    \item \texttt{epibench/t3/}: 151 pharmacogenomic
    MCQs with CPIC/ILAE gold standards.
    \item \texttt{epibench/t4/}: epilepsy-filtered
    questions from MedQA-USMLE (200) and
    MMLU Professional Medicine (272), with
    guideline-aligned gold standard answers.
    \item \texttt{epibench/t5/}: 163 PMC paper
    instances with expert annotations (30) and
    LLM-as-Judge gold standards (133).
    \item \texttt{scripts/}: retriever
    implementation (PPR-PCST, Semantic, Hybrid),
    evaluation scripts for all five tasks, and
    prompt templates.
\end{itemize}

\subsection{Author Statement}
\label{sec:app_doc_author}

We confirm that all data sources used in
\textsc{EpiKG} and \textsc{EpiBench} are publicly
available and used in accordance with their
respective licenses. We bear full responsibility
for ensuring compliance with license terms. All
ontology sources (ILAE, MeSH, HPO, OMIM, HGNC,
ChEBI, AES 2024) are credited with their original
citations in the main paper. The Harvard EEG
dataset is used under its stated academic use
terms~\cite{zafar2025harvard, sun2025harvard}.
MedQA-USMLE~\cite{jin2021disease} and MMLU
Professional Medicine~\cite{hendrycks2020measuring}
are used under their respective open licenses.
PMC papers used in T5 are accessed via the PubMed
Central Open Access Subset under CC-BY or CC0
licenses.

\subsection{Hosting, Licensing, and Maintenance}
\label{sec:app_doc_hosting}

\noindent\textbf{Hosting.}
\textsc{EpiKG}, \textsc{EpiBench}, code, and
project-page assets are hosted in the public
Hugging Face repository
\href{https://huggingface.co/RAI-Lab/EpiGraph}{\texttt{RAI-Lab/EpiGraph}}
and mirrored on GitHub at
\href{https://github.com/LabRAI/EpiGraph}{\texttt{LabRAI/EpiGraph}}.
Both platforms provide version control, allowing
users to track changes and ensure reproducibility
across experiments.

\noindent\textbf{Licensing.}
\textsc{EpiKG} and \textsc{EpiBench} are released
under the \textbf{CC BY 4.0} license, permitting
unrestricted academic use with attribution.
Downstream users are responsible for complying
with the licenses of individual source ontologies
and datasets.

\noindent\textbf{Maintenance.}
We commit to maintaining both resources for a
minimum of three years following publication.
Planned updates include: (1) annual \textsc{EpiKG}
refresh incorporating new epilepsy literature and
updated clinical guidelines; (2) expansion of
\textsc{EpiBench} T3 with new CPIC guideline
releases; and (3) community contributions via
GitHub pull requests for additional task
instances and evaluation metrics.

\subsection{Access and Reproducibility}
\label{sec:app_doc_access}

Users can access \textsc{EpiKG} and load it
directly via the Hugging Face Datasets library:

\begin{verbatim}
from datasets import load_dataset
epikg = load_dataset(
    "json",
    data_files="https://huggingface.co/RAI-Lab/EpiGraph/resolve/main/datasets/EpiKG/triplets.json",
    split="train",
)
\end{verbatim}

The \textsc{EpiBench} evaluation pipeline can
be run as follows:

\begin{verbatim}
git clone https://huggingface.co/RAI-Lab/EpiGraph
cd EpiGraph
pip install -r requirements.txt
python tasks/t1_clinical_decision_accuracy.py \
    --dataset datasets/EpiBench/t1_clinical_decision_accuracy_mcq.json \
    --triplets datasets/EpiKG/triplets.json \
    --model gpt-4o \
    --mode graph_rag
\end{verbatim}

Full environment specifications and
hyperparameter settings are provided in
\texttt{configs/} and documented in
Appendix~\ref{sec:app_retriever_hyper}.

\subsection{EpiKG Statistics}
\label{sec:app_doc_stats}

Table~\ref{tab:app_doc_stats} provides a
detailed breakdown of \textsc{EpiKG} entity
and relation statistics, complementing the
summary in Appendix~\ref{sec:app_epikg_stats}.

\begin{table}[h]
\centering
\small
\setlength{\tabcolsep}{5pt}
\renewcommand{\arraystretch}{1.2}
\caption{Detailed \textsc{EpiKG} statistics.
Cross-layer triplets cover all pairwise layer
combinations; the densest connections are between
L1 Syndrome and L4 Treatment (3,217 triplets)
and between L3 Gene and L1 Syndrome
(2,845 triplets).}
\label{tab:app_doc_stats}
\begin{tabular}{lrr}
\toprule
\textbf{Statistic} & \textbf{Value} & \textbf{(\%)} \\
\midrule
Total nodes          & 24,324 & 100.0 \\
Total edges          & 32,009 & 100.0 \\
\midrule
Rule-based triplets  & 9,670  & 30.2  \\
LLM-based triplets   & 22,339 & 69.8  \\
\midrule
Cross-layer triplets & 14,576 & 45.5  \\
Within-layer triplets & 17,433 & 54.5  \\
\midrule
Median paper count $\mathcal{P}$ & 3 & --- \\
IQR paper count      & 1--8   & --- \\
Triplets with $\mathcal{P} \geq 10$ & 4,612 & 14.4 \\
\bottomrule
\end{tabular}
\end{table}

\clearpage

\section{Prompts}
\label{sec:app_prompts}

\noindent\textbf{System prompt (all tasks).}
\begin{quote}
\ttfamily\small
You are an expert epileptologist with deep knowledge
of epilepsy syndromes, antiseizure medications,
pharmacogenomics, EEG interpretation, and epilepsy
research. Answer the following question based on the
provided clinical context and knowledge graph
evidence. Think step by step and ground your answer
in the evidence provided.
\end{quote}

\noindent\textbf{T1 Clinical Decision Accuracy MCQ input prompt.}
\begin{quote}
\ttfamily\small
Context: \{retrieved\_kg\_paths\}\\
Question: \{question\}\\
Options: A) \{opt\_a\} B) \{opt\_b\} C) \{opt\_c\}
D) \{opt\_d\}\\
Answer with the option letter and a brief
justification.
\end{quote}

\noindent\textbf{T1 Clinical Decision Accuracy Open-ended QA input prompt.}
\begin{quote}
\ttfamily\small
Context: \{retrieved\_kg\_paths\}\\
Question: \{question\}\\
Provide a detailed answer grounded in the provided
evidence. Cite specific entities or relations from
the knowledge graph context where relevant.
\end{quote}

\noindent\textbf{T2 Clinical Report Generation input prompt.}
\begin{quote}
\ttfamily\small
Context: \{retrieved\_kg\_paths\}\\
Patient history: \{patient\_history\}\\
EEG description: \{eeg\_description\}\\
Band power: \{band\_power\_stats\}\\
Spike rate: \{spike\_rate\_stats\}\\
Generate a clinical impression for this EEG report.
Your impression should include: (1) identification
of any epileptiform activity or abnormal patterns,
(2) syndrome or diagnosis consistent with the
findings, and (3) relevant clinical recommendations.
Ground your impression in the provided knowledge
graph evidence linking EEG patterns to syndromes
and treatments.
\end{quote}

\noindent\textbf{T3 Biomarker-Driven Precision Medicine input prompt.}
\begin{quote}
\ttfamily\small
Context: \{retrieved\_kg\_paths\}\\
Patient: \{genetic\_variant\}, \{phenotype\}\\
Select the most appropriate ASM from the following
options and justify your selection based on the
genetic evidence and clinical guidelines.\\
Options: A) \{opt\_a\} B) \{opt\_b\} C) \{opt\_c\}
D) \{opt\_d\}
\end{quote}

\noindent\textbf{T4 Treatment Recommendation input prompt.}
\begin{quote}
\ttfamily\small
Context: \{retrieved\_kg\_paths\}\\
Clinical scenario: \{clinical\_scenario\}\\
Select the most appropriate treatment option from
the following choices. Consider guideline
concordance, drug safety, and potential
contraindications based on the provided knowledge
graph evidence.\\
Options: A) \{opt\_a\} B) \{opt\_b\} C) \{opt\_c\}
D) \{opt\_d\}\\
Answer with the option letter and justify your
selection with reference to clinical guidelines
and any contraindication evidence.
\end{quote}

\noindent\textbf{T5 Deep Research Planning input prompt.}
\begin{quote}
\ttfamily\small
Context: \{retrieved\_kg\_paths\}\\
Paper: \{paper\_abstract\}\\
Generate a structured research plan covering:
(1) a focused research question,
(2) a study design rationale, and
(3) required data sources.
Ground your plan in the provided knowledge graph
evidence and identify at least one knowledge gap
not addressed by the paper.
\end{quote}

\clearpage

\section{Additional Experimental Results}
\label{sec:app_results}

\noindent\textbf{MMLU Treatment Recommendation.}
Table~\ref{tab:app_mmlu} reports full results on
the MMLU Professional Medicine subset, complementing
the MedQA-USMLE results in the main text.

\begin{table}[h]
\centering
\small
\setlength{\tabcolsep}{4pt}
\renewcommand{\arraystretch}{1.2}
\caption{T4 Treatment Recommendation: full MMLU
Professional Medicine results. DFS: Drug Safety
Score; GC: Guideline Concordance; KGEC: KG
Evidence Coverage. Results are mean $\pm$ std;
$\Delta$: relative improvement (\%).}
\label{tab:app_mmlu}

\begin{tabular}{llccccc}
\toprule
\textbf{Model} & \textbf{RAG} &
\textbf{Acc} & \textbf{$\Delta$} &
\textbf{DFS} & \textbf{GC} & \textbf{KGEC} \\
\midrule
\textit{GPT-4o}   & -- & 74.0$\pm$1.8 & -- & 0.35$\pm$0.03 & 70.0$\pm$2.0 & 0.66$\pm$0.02 \\
\rowcolor{blue!12}
& w/ & \textbf{83.0$\pm$1.5} & +12.2 & 0.44$\pm$0.02 & 78.0$\pm$1.7 & 0.72$\pm$0.02 \\
\textit{Claude S4} & -- & 71.0$\pm$1.9 & -- & 0.39$\pm$0.03 & 67.0$\pm$2.1 & 0.68$\pm$0.02 \\
\rowcolor{blue!12}
& w/ & 80.0$\pm$1.6 & +12.7 & \textbf{0.52$\pm$0.02} & \textbf{81.0$\pm$1.6} & \textbf{0.76$\pm$0.01} \\
\textit{Gemini}   & -- & 67.0$\pm$2.0 & -- & 0.36$\pm$0.03 & 62.0$\pm$2.3 & 0.62$\pm$0.03 \\
\rowcolor{blue!12}
& w/ & 76.0$\pm$1.7 & +13.4 & 0.43$\pm$0.02 & 74.0$\pm$1.9 & 0.71$\pm$0.02 \\
\textit{Llama}    & -- & 56.0$\pm$2.3 & -- & 0.30$\pm$0.04 & 53.0$\pm$2.5 & 0.63$\pm$0.03 \\
\rowcolor{blue!12}
& w/ & 68.0$\pm$1.9 & +21.4 & 0.42$\pm$0.03 & 66.0$\pm$2.1 & 0.69$\pm$0.02 \\
\textit{Qwen}     & -- & 54.0$\pm$2.4 & -- & 0.31$\pm$0.04 & 51.0$\pm$2.6 & 0.61$\pm$0.03 \\
\rowcolor{blue!12}
& w/ & 65.0$\pm$2.0 & +20.4 & 0.39$\pm$0.03 & 63.0$\pm$2.2 & 0.66$\pm$0.02 \\
\textit{Mistral}  & -- & 50.0$\pm$2.6 & -- & 0.29$\pm$0.04 & 47.0$\pm$2.8 & 0.58$\pm$0.03 \\
\rowcolor{blue!12}
& w/ & 61.0$\pm$2.1 & +22.0 & 0.36$\pm$0.03 & 59.0$\pm$2.3 & 0.64$\pm$0.02 \\
\midrule
\multicolumn{2}{l}{\textbf{Avg.\ $\Delta$}} &
\textbf{+17.2} & & \textbf{+28.1} & \textbf{+28.4} & \textbf{+13.2} \\
\bottomrule
\end{tabular}
\end{table}

\clearpage

\begin{figure*}[t]
    \centering
    \includegraphics[width=\linewidth]{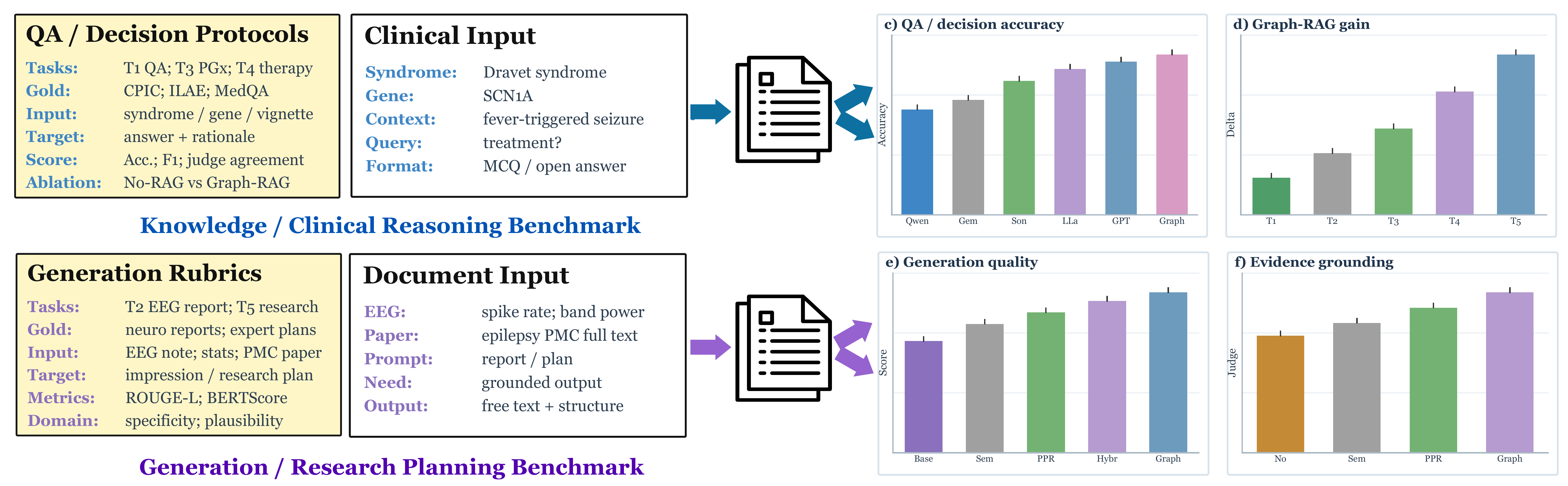}
    \caption{Overview of the \textbf{\textsc{EpiBench}}.}
    \label{fig:EpiBench_overview}
        \vspace{-0.3cm}
\end{figure*}

\begin{figure*}[t]
    \centering
    \includegraphics[width=\textwidth]{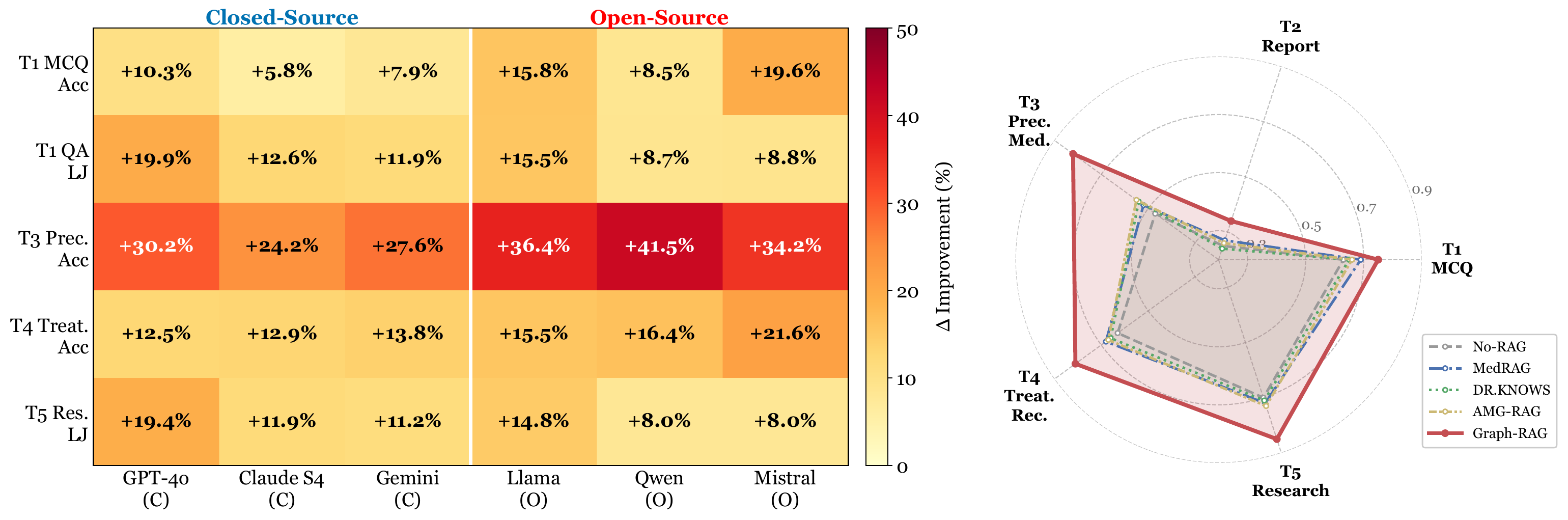}
            \vspace{-0.3cm}
    \caption{Overview of \textbf{\textsc{EpiBench}} Result.}
    \label{fig:overview_result}
\end{figure*}

\section{Limitations and Future Work}
\label{sec:app_limitations}

\noindent\textbf{Language and ontology coverage.}
\textsc{EpiKG} is constructed from English-language
ontologies and literature, limiting its coverage
of epilepsy syndromes and gene--disease associations
that are primarily documented in non-English sources.
Rare syndromes with fewer than five supporting
papers are likely underrepresented in the extracted
relation set, as the LLM-based extraction pipeline
requires sufficient co-occurrence evidence to produce
reliable triplets. Extending \textsc{EpiKG} to
multilingual sources and rare disease registries
is a natural direction for future work.

\noindent\textbf{LLM-as-Judge reliability.}
For T5 research planning, 133 of 163 gold standards
are produced by LLM-as-Judge rather than human
experts. While the judge prompt is designed to
minimise positional bias and hallucination, LLM
judges may exhibit systematic biases toward fluent
outputs regardless of scientific quality. Expanding
the expert-annotated subset and measuring
inter-rater reliability between human and LLM judges
is an important direction for strengthening the T5
evaluation.


\newpage

\end{document}